\documentclass{article}

\usepackage{microtype}
\usepackage{graphicx}
\usepackage{booktabs} 

\usepackage{hyperref}

\usepackage[accepted]{icml2020}

\usepackage{url}
\usepackage{mathtools}
\usepackage{amssymb,amsmath}
\usepackage{amsthm}
\usepackage{adjustbox}
\usepackage{cleveref}
\Crefname{equation}{Eq.}{Eqs.}
\usepackage{bbm}
\usepackage{stmaryrd}
\usepackage{subcaption}
\usepackage{natbib}
\usepackage{./notations}

\usepackage{dsfont}
\usepackage{enumerate}

\icmltitlerunning{Missing Data Imputation using Optimal Transport}

\begin{document}

\twocolumn[
\icmltitle{Missing Data Imputation using Optimal Transport}



\icmlsetsymbol{visit}{*}

\begin{icmlauthorlist}
\icmlauthor{Boris Muzellec}{ensae}
\icmlauthor{Julie Josse}{inria,cmap,visit}
\icmlauthor{Claire Boyer}{lpsm}
\icmlauthor{Marco Cuturi}{goo,ensae}
\end{icmlauthorlist}

\icmlaffiliation{ensae}{CREST-ENSAE, IP Paris, Palaiseau, France}
\icmlaffiliation{inria}{XPOP, INRIA Saclay, France}
\icmlaffiliation{goo}{Google Brain, Paris, France}
\icmlaffiliation{cmap}{CMAP, UMR7641, \'{E}cole Polytechnique, IP Paris, Palaiseau, France}
\icmlaffiliation{lpsm}{LPSM, Sorbonne Universit\'{e}, ENS Paris, France}

\icmlcorrespondingauthor{Boris Muzellec}{boris.muzellec@ensae.fr}

\icmlkeywords{}

\vskip 0.3in
]

\printAffiliationsAndNotice{This research was done while JJ was a visiting researcher at Google Brain Paris.} 

\begin{abstract}
Missing data is a crucial issue when applying machine learning algorithms to real-world datasets. Starting from the simple assumption that two batches extracted randomly from the same dataset should share the same distribution, we leverage optimal transport distances to quantify that criterion and turn it into a loss function to impute missing data values. We propose practical methods to minimize these losses using end-to-end learning, that can exploit or not parametric assumptions on the underlying distributions of values. We evaluate our methods on datasets from the UCI repository, in MCAR, MAR and MNAR settings. These experiments show that OT-based methods match or out-perform state-of-the-art imputation methods, even for high percentages of missing values.
\end{abstract}

\section{Introduction}\label{sec:intro}
Data collection is usually a messy process, resulting in datasets that have many missing values. This has been an issue for as long as data scientists have prepared, curated and obtained data, and is all the more inevitable given the vast amounts of data currently collected.
The literature on the subject is therefore abundant \citep{little2002statistical,buuren_2018}: a recent survey indicates that there are more than 150 implementations available to handle missing data \citep{mayer2019}.
These methods differ on the objectives of their analysis (estimation of parameters and their variance, matrix completion, prediction), the nature of the variables considered (categorical, mixed, etc.), the assumptions about the data, and the missing data mechanisms. 
Imputation methods, which consist in filling missing entries with plausible values, are very appealing as they allow to both get a guess for the missing entries as well as to perform (with care) downstream machine learning methods on the completed data. Efficient methods include, among others, methods based on low-rank assumptions \citep{hastie2015softimpute}, iterative random forests \citep{Stekhoven2012forest} and imputation using variational autoencoders \citep{mattei19a, ivanov2018variational}. 
A desirable property for imputation methods is that they should preserve the joint and marginal distributions of the data. Non-parametric Bayesian strategies \citep{murray2016} or recent approaches based on generative adversarial networks \citep{yoon18a} are attempts in this direction. However, they can be quite cumbersome to implement in practice. 

We argue in this work that the optimal transport (OT) toolbox constitutes a natural, sound and straightforward alternative. Indeed, optimal transport provides geometrically meaningful distances to compare discrete distributions, and therefore data. 
Furthermore, thanks to recent computational advances grounded on regularization~\citep{cuturi2013sinkhorn}, OT-based divergences can be computed in a scalable and differentiable way~\citep{peyre2019computational}. Those advances have allowed to successfully use OT as a loss function in many applications, including multi-label classification~\citep{frogner2015learning}, inference of pathways~\cite{schiebinger2019optimal} and generative modeling~\citep{arjovsky2017wgan,genevay18a,salimans2018improving}. Considering the similarities between generative modeling and missing data imputation, it is therefore quite natural to use OT as a loss for the latter.

\textbf{Contributions.}
This paper presents two main contributions. First, we leverage OT to define a loss function for missing value imputation. This loss function is the mathematical translation of the simple intuition that two random batches from the same dataset should follow the same distribution. Next, we provide algorithms for imputing missing values according to this loss. Two types of algorithms are presented, the first (i) being non-parametric, and the second (ii) defining a class of parametric models. 
The non-parametric algorithm (i) enjoys the most degrees of freedom, and can therefore output imputations which respect the global shape of the data while taking into account its local features. 
The parametric algorithm (ii) is trained in a round-robin fashion similar to iterative conditional imputation techniques, as implemented for instance in the \texttt{mice} package \cite{buuren2011}. Compared to the non-parametric method, this algorithm allows to perform out-of-sample imputation. This creates a very flexible framework which can be combined with many imputing strategies, including imputation with Multi-Layer Perceptrons.
Finally, these methods are showcased in extensive experiments on a variety of datasets and for different missing values proportions and mechanisms, including the difficult case of informative missing entries. The code to reproduce these experiments is available at \url{https://github.com/BorisMuzellec/MissingDataOT}.

\textbf{Notations.}
Let $\mask =(\maskij_{ij})_{ij} \in \{0,1\}^{n \times \dim}$ be a binary mask encoding observed entries, i.e.\ $\maskij_{ij}=1$ (resp. $0$) \textit{iff} the entry $(i,j)$ is observed (resp. missing). We observe the following incomplete data matrix:
$$
\data = \Xobs \odot \mask + {\tt NA}\odot (\mathds{1}_{n\times \dim}-\mask),
$$
where $\Xobs\in \RR^{n\times \dim}$ contains the observed entries, $\odot$ is the elementwise product and $\mathds{1}_{n\times \dim}$ is an $n\times \dim$ matrix filled with ones. Given the data matrix $\data$, our goal is to construct an estimate $\hat{\data}$ filling the missing entries of $\data$, which can be written as 
$$
\hat{\data} = \Xobs \odot \mask + {\hXmis}\odot (\mathds{1}_{n\times \dim}-\mask),
$$
where  ${\hXmis}\in \RR^{n\times \dim}$ contains the imputed values. Let $\datarow_{i:}$ denote the $i$-th row of the data set $\data$, such that $\data = (\datarow_{i:}^T)_{1\leq i\leq n}$. Similarly, $\datarow_{:j}$ denotes the $j$-th column (variable) of the data set $\data$, such that $\data = (\datarow_{:1} | \hdots | \datarow_{:\dim})$, and $\data_{:-j}$ denotes the dataset $\data$ in which the $j$-th variable has been removed. For $K\subset \{1,\hdots , n\}$ a set of $m$ indices, $\data_K = (\datarow_{k:})_{k\in K}$ denotes the corresponding batch, and by $\mu_m(\data_K)$ the empirical measure associated to $\data_K$, i.e.\ 
$$
\mu_m(\data_K) := \tfrac{1}{m}\sum_{k\in K} \delta_{\datarow_{k:}}.
$$
Finally, $\Delta_n \defeq \{\ba\in\RR^n_+ : \sum_{i=1}^n a_i = 1\}$ is the simplex in dimension $n$.

\section{Background}\label{sec:background}\BM{R1: too long compared to section 3}

\textbf{Missing data}. 
\citet{rubin1976inference} defined a widely used - yet controversial \citep{seaman2013meant} -  nomenclature for missing values mechanisms.
This nomenclature distinguishes between three cases: missing completely at random (MCAR), missing at random (MAR), and missing not at random (MNAR). In MCAR, the missingness is independent of the data, whereas in MAR, the probability of being missing depends only on observed values. A subsequent part of the literature, with notable exceptions \citep{kim2018data, mohan:pea19-r473}, only consider these 
``simple'' mechanisms and struggles for the harder yet prevalent MNAR case. MNAR values lead to important biases in the data, as the probability of missingness then depends on the unobserved values.
On the other hand, MCAR and MAR are ``ignorable'' mechanisms in the sense that they do not make it necessary to model explicitly the distribution of missing values when maximizing the observed likelihood. 

The naive workaround which consists in deleting observations with missing entries is not an alternative in high dimension. Indeed, let us assume as in \citet{zhu2019high}
that $\data$ is a $n \times \dim$ data matrix in which each
entry is missing independently with probability 0.01. When $\dim = 5$, this would result in around 95\% of the individuals (rows) being retained, but for $\dim = 300$, only around 5\% of rows have no missing entries. Hence, providing plausible imputations for missing values quickly becomes necessary.
Classical imputation methods impute according to a joint distribution which is either explicit, or implicitly defined through a set of conditional distributions. 
As an example, explicit joint modeling methods include imputation models that assume a Gaussian distribution for the data, whose parameters are estimated using EM algorithms \citep{dempster1977maximum}. Missing values are then imputed by drawing from their predictive distribution. A second instance of such joint modeling methods are imputations assuming low-rank structure \citep{josse2016missmda}. 
The conditional modeling approach \citep{buuren_2018}, also known as ``sequential imputation'' or ``imputation using chained equations'' (ice) consists
in specifying one model for each variable. It predicts the missing values of each variable using the other variables as explanatory, and cycles through the variables iterating this procedure to update the imputations until predictions stabilize. 

Non-parametric methods like $k$-nearest neighbors imputation \citep{troyanskaya2001missing} or random forest imputation \citep{Stekhoven2012forest} have also been developed and account for the local geometry of the data. The herein proposed methods lie at the intersection of global and local approaches and are derived in a non-parametric and parametric version. 

\textbf{Wasserstein distances, entropic regularization and Sinkhorn divergences.}\BM{R5: Better introduction to OT in general before jumping to Sinkhorn etc.}
Let $\alpha = \sum_{i=1}^n a_i\delta_{\bx_i}$, $\beta = \sum_{i=1}^{n'} b_i\delta_{\by_i}$ be two discrete distributions, described by their supports $(\bx_i)_{i=1}^n \in \RR^{n\times p}$ and  $(\by_i)_{i=1}^{n'} \in \RR^{n'\times p}$ and weight vectors $\ba \in \Delta_n$ and $\bb\in \Delta_{n'}$. Optimal transport compares $\alpha$ and $\beta$ by considering the most efficient of transporting the masses $\ba$ and $\bb$ onto each-other, according to a ground cost between the supports. The (2-)Wasserstein distance corresponds to the case where this ground cost is quadratic:
\begin{equation}\label{eq:W2}
    W_2^2(\alpha, \beta) \defeq \underset{\bP \in U(\ba, \bb)}{\min}\dotp{\bP}{\bM},
\end{equation}
where $U(\ba, \bb) \defeq \{\bP\in \RR^{n\times n'} : \bP\mathds{1}_{n'} = \ba, \bP^T\mathds{1}_n = \bb\}$ is the set of transportation plans, and $\bM = \left( \|x_i - y_j\|^2 \right)_{ij} \in \RR^{n\times n'}$ is the matrix of pairwise squared distances between the supports. $W_2$ is not differentiable and requires solving a costly linear program via network simplex methods~\citep[\S3]{peyre2019computational}. Entropic regularization alleviates both issues: consider
\begin{equation}\label{eq:ot_reg}
    \OT_\varepsilon(\alpha, \beta) \defeq \underset{\bP \in U(\ba, \bb)}{\min}\dotp{\bP}{\bM} + \varepsilon h(\bP),
\end{equation}
where $\varepsilon > 0$ and $h(\bP) \defeq \sum_{ij} p_{ij}\log p_{ij}$ is the negative entropy. Then, $\OT_\varepsilon(\alpha, \beta)$ is differentiable and can be solved using Sinkhorn iterations~\citep{cuturi2013sinkhorn}. However, due to the entropy term, $\OT_\varepsilon$ is no longer positive. This issue is solved through debiasing, by subtracting auto-correlation terms. Let 
\begin{equation}\label{eq:sk_div}
    S_\varepsilon(\alpha, \beta) \defeq \OT_\varepsilon(\alpha, \beta) - \frac{1}{2}(\OT_\varepsilon(\alpha, \alpha) + \OT_\varepsilon(\beta, \beta)).
\end{equation}
\Cref{eq:sk_div} defines the Sinkhorn divergences~\citep{genevay18a}, which are positive, convex, and can be computed with little additional cost compared to entropic OT~\citep{feydy2019}. Sinkhorn divergences hence provide a differentiable and tractable proxy for Wasserstein distances, and will be used in the following.

\textbf{OT gradient-based methods.}
Not only are the OT metrics described above good measures of distributional closeness, they are also well-adapted to gradient-based imputation methods. Indeed, let $\bX_K, \bX_L$ be two batches drawn from $\bX$. Then, gradient updates for $\OT_\varepsilon(\mu_m(\bX_K), \mu_m(\bX_L)), \varepsilon \geq 0$ w.r.t a point $\datarow_{k:}$ in $\bX_K$ correspond to taking steps along the so-called barycentric transport map. Indeed, with (half) quadratic costs, it holds \citep[\S4.3]{CuturiBarycenter} that
$$ \nabla_{\datarow_{k:}} \OT_\varepsilon(\mu_m(\bX_K), \mu_m(\bX_L)) = \sum_{\ell} \bP^\star_{k\ell}  (\datarow_{k:} - \datarow_{\ell:}),$$
where $\bP^\star$ is the optimal (regularized) transport plan. Therefore, a gradient based-update is of the form 
\begin{equation}\label{eq:OT_grad_update}
\datarow_{k:} \leftarrow (1 - t) \datarow_{k:} + t \sum_{l} \bP^\star_{kl} \datarow_{l:}.
\end{equation}
In a missing value imputation context, \Cref{eq:OT_grad_update} thus corresponds to updating values to make them closer to the target points given by transportation plans. Building on this fact, OT gradient-based imputation methods are proposed in the next section.
\section{Imputing Missing Values using OT}\label{sec:miss_data_ot}
Let $\data_K$ and $\data_L$ be two batches respectively extracted from the complete rows and the incomplete rows in $\data$, such that only the batch $\data_L$ contains missing values. 
In this one-sided incomplete batch setting, a good imputation should preserve the distribution from the complete batch, meaning that $\data_K$ should be close to $\data_L$ in terms of distributions. The OT-based metrics described in \Cref{sec:background} provide natural criteria to catch this distributional proximity and derive imputation methods.
However, as observed in \Cref{sec:background}, in high dimension or with a high proportion of missing values, it is unlikely or even impossible to obtain batches from $\data$ with no missing values. Nonetheless, a good imputation method should still ensure that the distributions of any two i.i.d.\ {incomplete} batches $\data_K$ and $\data_{L}$, \emph{both} containing missing values, should be close. This implies in particular that OT-metrics between the distributions $\mu_m{\data_K}$ and $\mu_m{\data_L}$ should have small values. This criterion, which is weaker than the one above with one-sided missing data but is more amenable, will be considered from now on.

\textbf{Direct imputation.} \Cref{alg:sinkhorn_imp} is a direct implementation of this criterion, aiming to impute missing values for quantitative variables by minimizing OT distances between batches. 
First, missing values of any variable are initialized with the mean of observed entries plus a small amount of noise (to preserve the marginals and to facilitate the optimization). Then, batches are sequentially sampled and the Sinkhorn divergence between batches is minimized with respect to the imputed values, using gradient updates (here using RMSprop~\cite{tieleman2012lecture}).

\begin{algorithm}[ht]
\caption{Batch Sinkhorn Imputation}\label{alg:sinkhorn_imp}
\begin{algorithmic} 
\REQUIRE $\bX\in (\RR \cup \{\tt{NA}\} )^{n\times \dim}$, $\mask\in\{0,1\}^{n\times \dim}$, $\alpha, \eta, \varepsilon > 0$, $n\geq m > 0$, 
\STATE \textbf{Initialization}: for $j=1, \hdots, \dim$,
\begin{itemize}
    \item  for $i$ s.t. $\maskij_{ij}=0$, $\hat{x}_{ij} \leftarrow \overline{\bx^{obs}_{:j}} + \varepsilon_{ij},$
with $\varepsilon_{ij} \sim \Ncal(0, \eta)$ and $\overline{\bx^{obs}_{:j}}$ corresponds to the mean of the observed entries in the $j$-th variable (missing entries)
    \item for $i$ s.t. $\maskij_{ij}=1$, $\hat{x}_{ij} \leftarrow x_{ij}$ (observed entries)
\end{itemize}
\FOR{t = 1, 2, ...,$t_{max}$}
\STATE Sample two sets $K$ and $L$ of $m$ indices
\STATE $\Lcal (\hat{\data}_K, \hat{\data}_L) \leftarrow S_\varepsilon(\mu_m (\hat{\data}_K), \mu_m (\hat{\data}_L))$
\STATE $\hXmis_{K \cup L} \leftarrow \hXmis_{K \cup L} - \alpha\text{RMSprop}(\nabla_{\hXmis_{K \cup L}}\Lcal)$
\ENDFOR
\ENSURE $\hat{\data}$
\end{algorithmic}
\end{algorithm}

\textbf{OT as a loss for missing data imputation.}
Taking a step back, one can see that \Cref{alg:sinkhorn_imp} essentially uses Sinkhorn divergences between batches as a loss function to impute values for a model in which ``one parameter equals one imputed value''. Formally, for a fixed batch size $m$, this loss is defined as
\begin{align}\label{eq:loss_batches}
\Lcal_m(\bX) \defeq \sum_{\mathclap{\substack{K : 0 \leq k_1 < ... < k_m \leq n\\
                         L : 0 \leq \ell_1 < ... < \ell_m \leq n}}}  
                         S_\varepsilon(\mu_m(\data_K), \mu_m(\data_L)).
\end{align}
\Cref{eq:loss_batches} corresponds to the ``autocorrelation" 
counterpart of the minibatch Wasserstein distances described in~\citet{fatras2019,salimans2018improving}.

Although \Cref{alg:sinkhorn_imp} is straightforward, a downside is that it cannot directly generate imputations for out-of-sample data points with missing values. Hence, a natural extension is to use the loss defined in \Cref{eq:loss_batches} to fit parametric imputation models, provided they are differentiable with respect to their parameters. At a high level, this method is described by \Cref{alg:meta_sinkhorn}.
\begin{algorithm}[ht]
\caption{Meta Sinkhorn Imputation}\label{alg:meta_sinkhorn}
\begin{algorithmic} 
\REQUIRE $\bX\in\RR^{n\times \dim}$, $\mask\in\{0,1\}^{n\times \dim}$, $\text{Imputer}(\cdot, \cdot, \cdot)$, $\Theta_0$, $\varepsilon > 0$, $n\geq m > 0$, 
\STATE $\hat{\data}^0 \leftarrow$ same initialization as in \Cref{alg:sinkhorn_imp}
\STATE $\hat{\Theta} \leftarrow \Theta_0$
\FOR{$t = 1, 2, ..., t_{\text{max}}$}
\FOR{$k = 1, 2, ..., K$}
\STATE $\hat{\data} \leftarrow \text{Imputer}(\hat{\data}^t, \Omega, \hat{\Theta})$ 
\STATE Sample two sets $K$ and $L$ of $m$ indices 
\STATE $\Lcal(\hat{\data}_K, \hat{\data}_L) \leftarrow S_\varepsilon(\mu_m (\hat{\data}_K)  ,\mu_m (\hat{\data}_L))$
\STATE $\nabla_\Theta \Lcal \leftarrow \text{AutoDiff}(\Lcal(\hat{\data}_K, \hat{\data}_L))$
\STATE $\hat{\Theta} \leftarrow \hat{\Theta} - \alpha\text{Adam}(\nabla_\Theta \Lcal)$
\ENDFOR
\STATE $\hat{\bX}^{t+1} \leftarrow \text{Imputer}(\hat{\bX}^t, \Omega, \hat{\Theta})$  
\ENDFOR
\ENSURE Completed data $\hat{\bX} = \hat{\bX}^{t_{\text{max}}}$, $\text{Imputer}(\cdot,\cdot, \hat{\Theta})$
\end{algorithmic}
\end{algorithm}
\Cref{alg:meta_sinkhorn} takes as an input an imputer model with a parameter $\Theta$ such that $\text{Imputer}(\data,\Omega, \Theta)$ returns imputations for the missing values in $\data$. This imputer has to be differentiable w.r.t. its parameter $\Theta$, so that the batch Sinkhorn loss $\Lcal$ can be back-propagated through $\hat{\data}$ to perform gradient-based updates of $\Theta$. \Cref{alg:meta_sinkhorn} does not only return the completed data matrix $\hat{\data}$, but also the trained parameter $\hat{\Theta}$, which can then be re-used to impute missing values in out-of-sample data.

\textbf{Round-robin imputation.}
\begin{figure*}[ht]
    \centering
    \includegraphics[width=1.\linewidth]{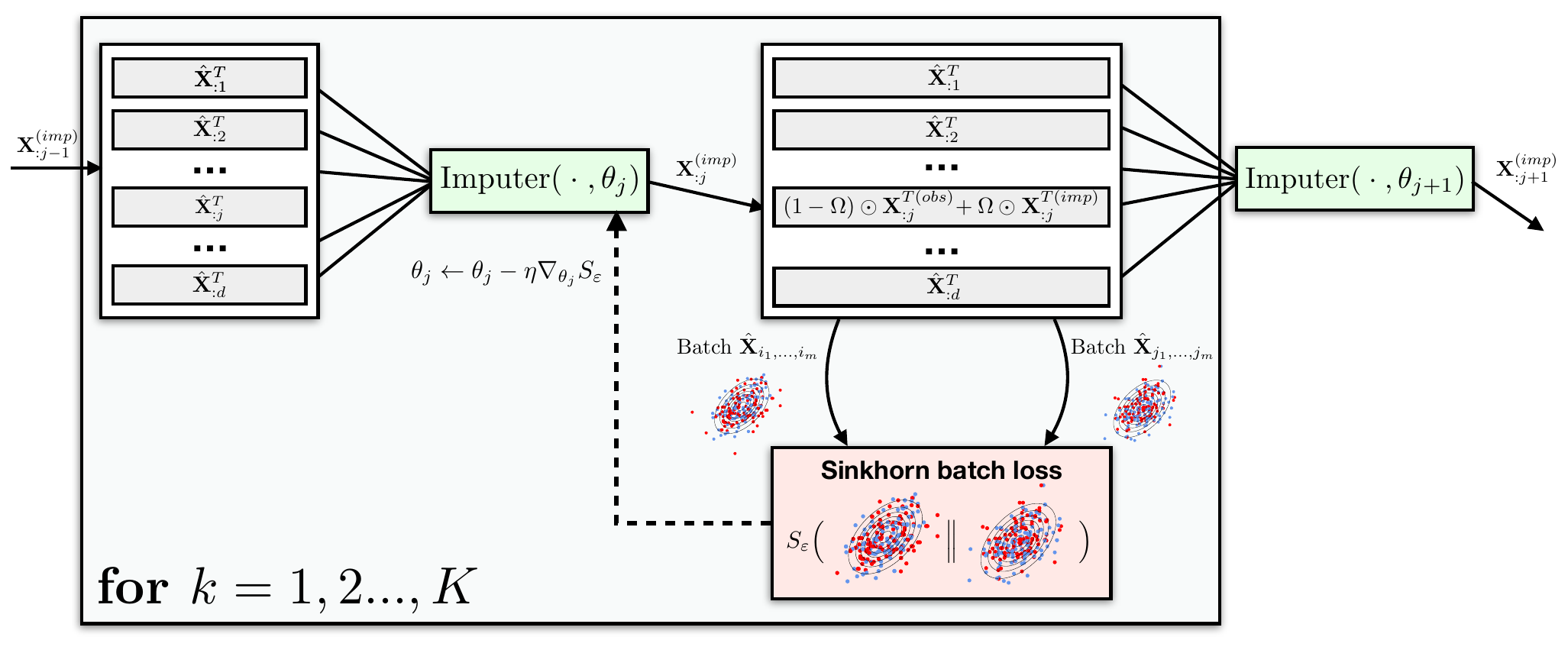}
    \vskip-.1cm
    \caption{Round-robin imputation: illustration of the imputation of the $j$-th variable in the inner-most loop of \Cref{alg:rr_sinkhorn}.}
    \label{fig:round_robin}
    \vskip-.3cm
\end{figure*}
A remaining unaddressed point in \Cref{alg:meta_sinkhorn} is how to perform the ``$\hat{\data} \leftarrow \text{Imputer}(\hat{\data}^t, \Omega, \Theta)$'' step in the presence of missing values. 
A classical method is to perform imputations over variables in a round-robin fashion, i.e. to iteratively predict missing coordinates using other coordinates as features in a cyclical manner. The main advantage of this method is that it decouples variables being used as inputs and those being imputed.
This requires having $\dim$ sets of parameter $(\theta_j)_{1\leq j \leq \dim}$, one for each variable, where each $\theta_j$ refers to the parameters used to to predict the $j$-th variable. 
The $j$-th variable is iteratively imputed using the $\dim-1$ remaining variables, according to the chosen model with parameter $\theta_j$: $\hat{\theta}_j$ is first fitted (using e.g. regression or Bayesian methods), then the $j$-th variable is imputed. The algorithm then moves to the next variable $j+1$, in a cyclical manner. This round-robin method is implemented for instance in \texttt{R}'s \texttt{mice} package~\citep{buuren2011} or in the \texttt{IterativeImputer} method of the \texttt{scikit-learn}~\citep{scikit-learn} package. 
When using the Sinkhorn batch loss \cref{eq:loss_batches} to fit the imputers, this procedure can be seen as a particular case of \Cref{alg:meta_sinkhorn} where the imputer parameter $\Theta$ is separable with respect to each variable $(\datarow_{:j})_{1\leq j \leq \dim}$, i.e.\ $\Theta$ consists in $\dim$ sets of parameter  $(\theta_j)_{1\leq j \leq \dim}$. 

Making this round-robin imputation explicit in the step ``$\hat{\data} \leftarrow \text{Imputer}(\hat{\data}^t, \Omega, \Theta)$'' of \Cref{alg:meta_sinkhorn} leads to \Cref{alg:rr_sinkhorn}.
\begin{algorithm}[ht]
\caption{Round-Robin Sinkhorn Imputation}\label{alg:rr_sinkhorn}
\begin{algorithmic} 
\REQUIRE $\data\in\RR^{n\times \dim}$, $\mask\in\{0,1\}^{n\times \dim}$, $\text{Imputer}(\cdot, \cdot, \cdot)$, $\Theta_0$, $\varepsilon > 0$, $n\geq m > 0$,
\STATE $\hat{\data}^0 \leftarrow$ same initialization as in \Cref{alg:sinkhorn_imp}
\STATE $(\hat{\theta}_1, ..., \hat{\theta}_\dim) \leftarrow \Theta_0$
\FOR{$t = 1, 2, ..., t_{\text{max}}$}
\FOR{$j = 1, 2, ..., \dim$}
\FOR{$k = 1, 2, ..., K$}
\STATE $\hat{\data}_{:j} \leftarrow \text{Imputer}(\hat{\data}_{:-j}^t, \Omega_{:j}, \hat{\theta}_j)$  
\STATE Sample two sets $K$ and $L$ of $m$ indices 
\STATE $\Lcal \leftarrow S_\varepsilon(\mu_m (\hat{\data}_K), \mu_m (\hat{\data}_L))$
\STATE $\nabla_{\theta_j}\Lcal \leftarrow \text{AutoDiff}(\Lcal)$
\STATE $\hat{\theta}_j \leftarrow \hat{\theta}_j - \alpha\text{Adam}(\nabla_{\theta_j} \Lcal)$
\ENDFOR
\STATE $\hat{\data}_{:j}^{t} \leftarrow \text{Imputer}(\hat{\data}_{:-j}^t,\Omega_{:j}, \hat{\theta}_j)$  
\ENDFOR
\STATE $\hat{\data}^{t+1} \leftarrow \hat{\data}^{t}$
\ENDFOR
\ENSURE Imputations $\hat{\data}^{t_{\text{max}}}$, $\text{Imputer}(\cdot, \cdot, \hat{\Theta})$
\end{algorithmic}
\end{algorithm}
In \Cref{alg:rr_sinkhorn}, an imputation $\hat{\data}^t, t = 0, ..., t_{\text{max}}$ is updated starting from an initial guess $\hat{\data}^0$. 
The algorithm then consists in three nested loops. (i) The inner-most loop is dedicated to gradient-based updates of the parameter $\hat{\theta}_j$, as illustrated in \Cref{fig:round_robin}.
Once this inner-most loop is finished, the $j$-th variable of $\hat{\bX}^t$ is updated using the last update of $\hat{\theta}_j$.
(ii) This is performed cyclically over all variables of $\hat{\bX}^t$, yielding $\hat{\data}^{t+1}$. (iii) This fitting-and-imputation procedure over all variables is repeated until convergence, or until a given number of iterations is reached. 

In practice, several improvements on the generic \Cref{alg:meta_sinkhorn,alg:rr_sinkhorn} can be implemented: 
\begin{enumerate}
    \item To better estimate \Cref{eq:loss_batches}, one can sample several pairs of batches (instead of a single one) and define $\Lcal$ as the average of $S_\varepsilon$ divergences.
    \item For \Cref{alg:rr_sinkhorn} in a MCAR setting, instead of sampling in each pair two batches from $\hat{\bX}$, one of the two batches can be sampled with no missing value on the $j$-th variable, and the other with missing values on the $j$-th variable. This allows the imputations for the $j$-th variable to be fitted on actual non-missing values. This helps ensuring that the imputations for the $j$-th variable will have a marginal distribution close to that of non-missing values.\label{enum:unsym_batches}
    \item The order in which the variables are imputed can be adapted. A simple heuristic is to impute variables in increasing order of missing values.
    \item During training, the loss can be hard to monitor due to the high variance induced by estimating \Cref{eq:loss_batches} from a few pairs of batches. Therefore, it can be useful to define a validation set on which fictional additional missing values are sampled to monitor the training of the algorithm, according to the desired accuracy  score (e.g. MAE, RMSE or $W_2$ as in \Cref{sec:exps}).
\end{enumerate}
Note that \cref{enum:unsym_batches} is \textit{a priori} only legitimate in a MCAR setting. Indeed, under MAR or MNAR assumptions, the distribution of non-missing data is in general not equal to the original (unknown) distribution of missing data.\footnote{Consider as an example census data in which low/high income people are more likely to fail to answer an income survey than medium income people.} Finally, the use of Adam~\citep{kingma2014adam} compared to RMSprop in \Cref{alg:sinkhorn_imp} is motivated by empirical performance, but does not have a crucial impact on performance. It was observed however that the quality of the imputations given by \Cref{alg:sinkhorn_imp} seems to decrease when gradient updates with momentum are used.

\section{Experimental Results}\label{sec:exps}
\begin{figure}[ht]
    \begin{subfigure}{1.\columnwidth}
      \centering
      \includegraphics[trim={0 0 0 0},clip,width=1.\linewidth]{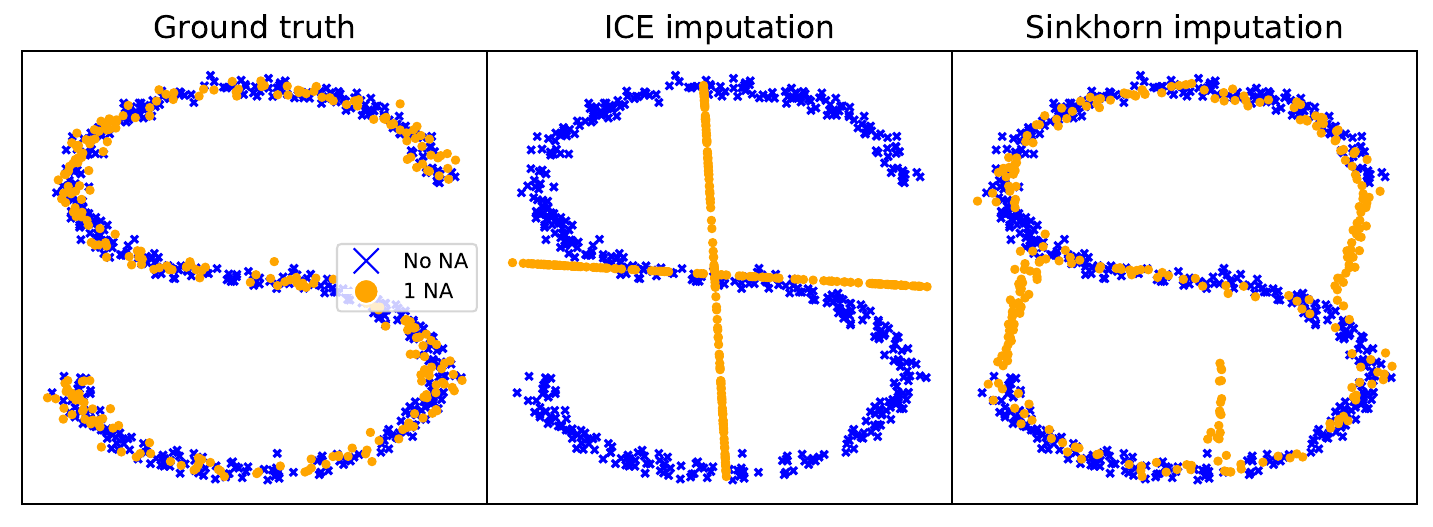}
    \end{subfigure}\hfill
      \begin{subfigure}{1.\columnwidth}
      \centering
      \includegraphics[trim={0 0 0 .85cm},clip,width=1.\linewidth]{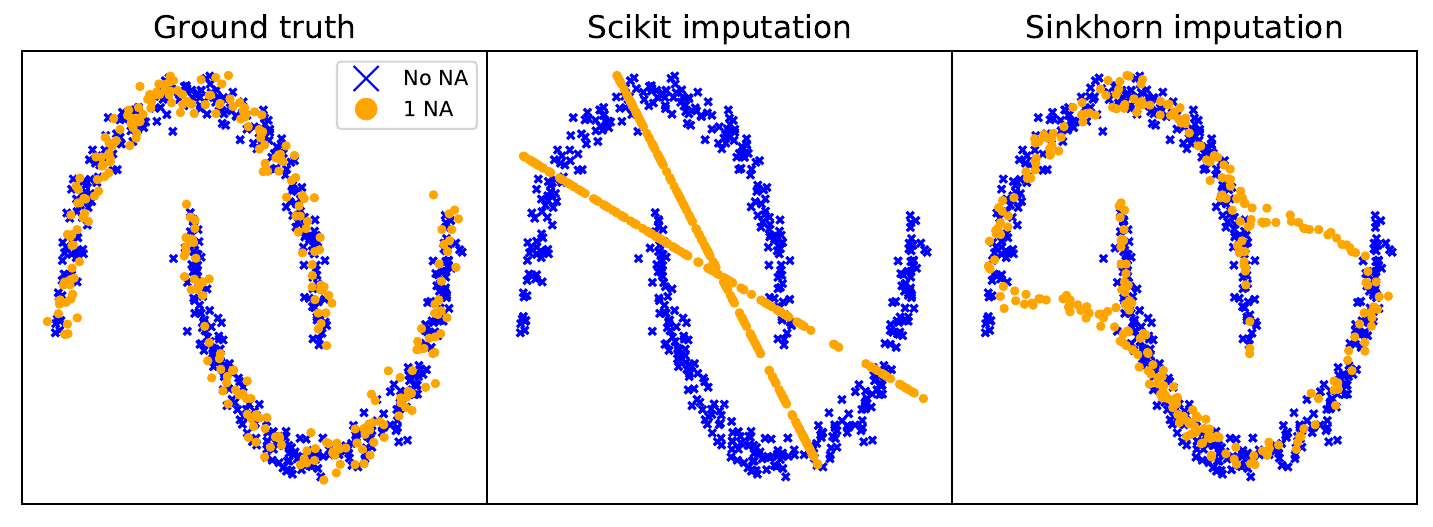}
    \end{subfigure}\hfill
    \begin{subfigure}{1.\columnwidth}
      \centering
      \includegraphics[trim={0 0 0 .85cm},clip,width=1.\linewidth]{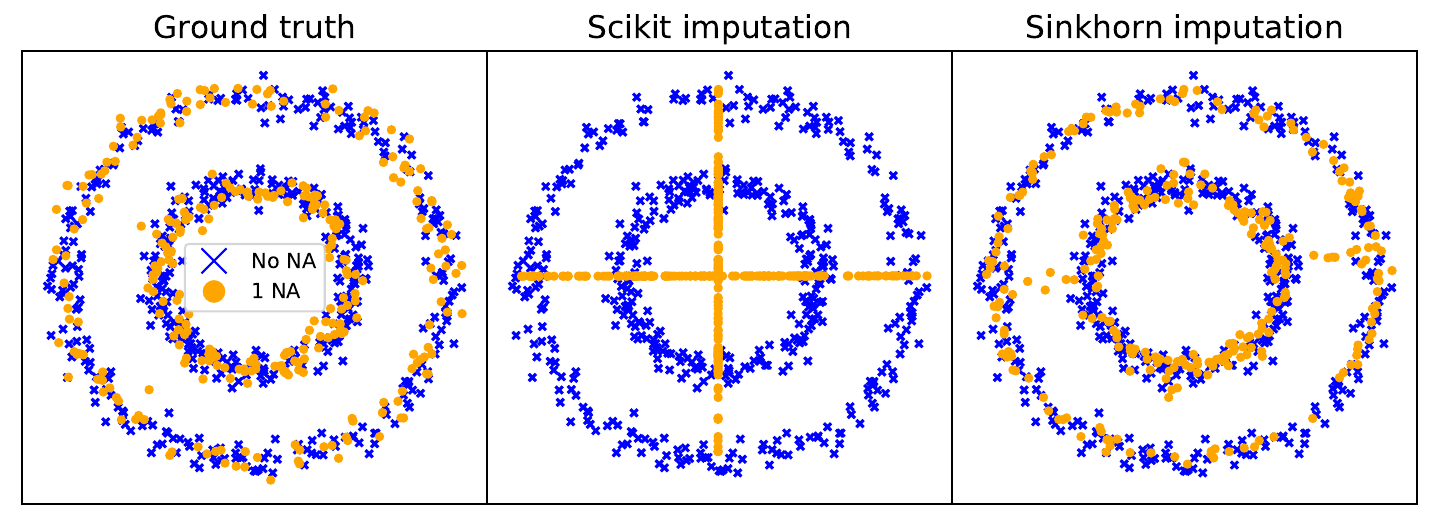}
    \end{subfigure}\hfill
    \begin{subfigure}{1.\columnwidth}
      \centering
      \includegraphics[trim={0.cm 0.cm 0.cm .85cm},clip,width=1.\linewidth]{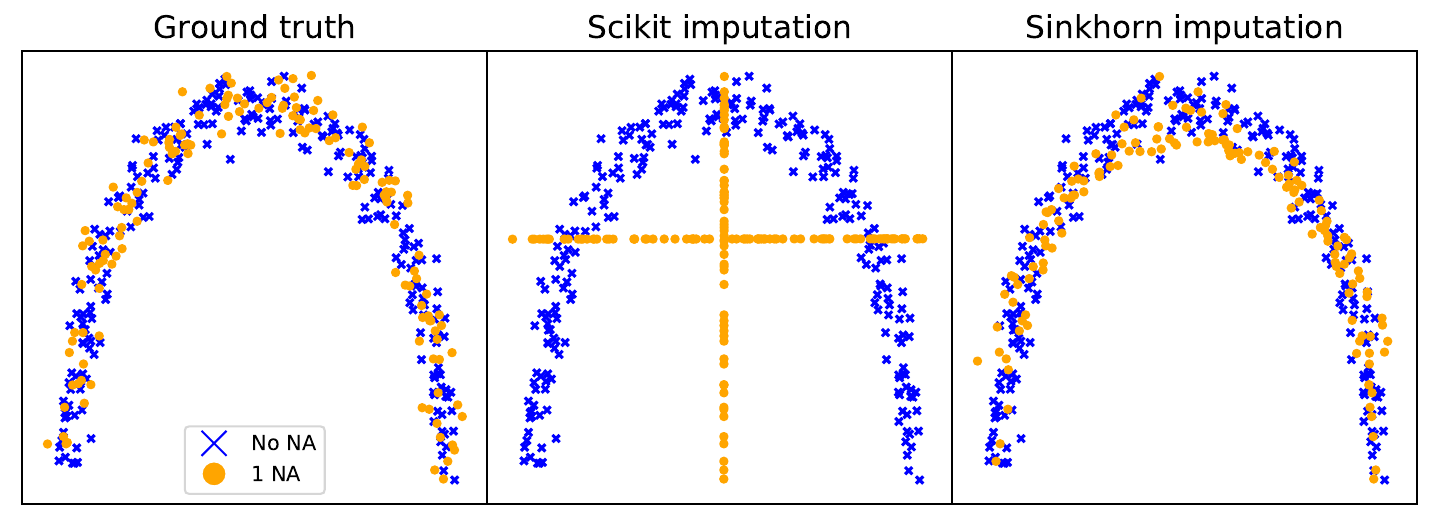}
    \end{subfigure}
    \vskip-.3cm
    \caption{Toy examples: 20 \% missing values (MCAR) on toy datasets. Blue points have no missing values, orange points have one missing value on either coordinate. \textbf{ice} outputs conditional expectation imputations, which are irrelevant due to the high non-linearity of these examples. Since \cref{alg:sinkhorn_imp} does not assume a parametric form for the imputations, it is able to satisfyingly impute missing values.}
    \label{fig:toy_example}
    \vskip-.3cm
\end{figure}
\begin{figure*}[ht]
    \centering
    \includegraphics[width = 1.\textwidth]{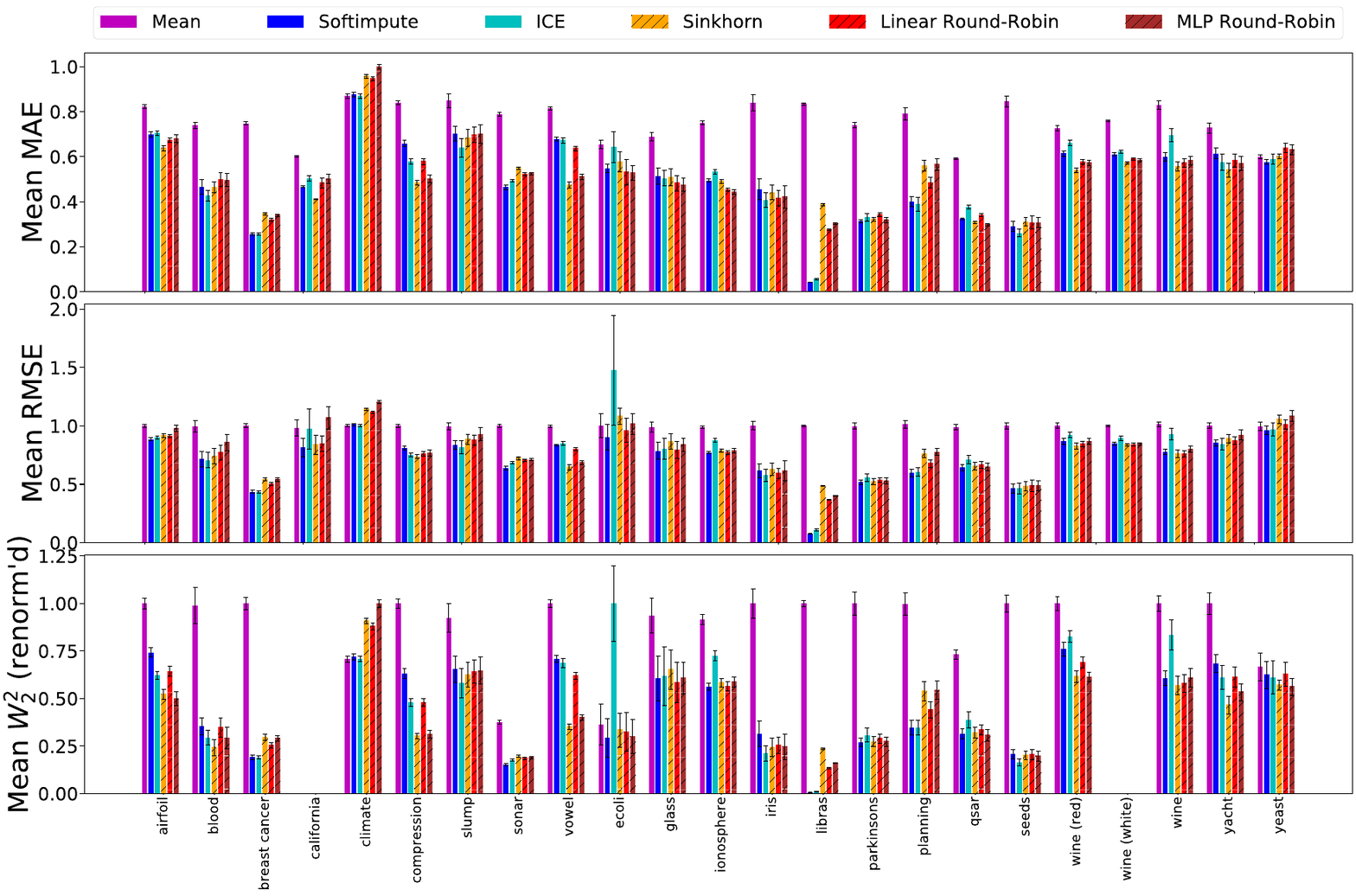}
    \vskip-.3cm
    \caption{(\textbf{30\% MCAR}) Imputation methods on 23 datasets from the UCI repository (\Cref{tab:datasets}). \texttt{Sinkhorn} denotes \Cref{alg:sinkhorn_imp} and \texttt{Linear RR},\texttt{ MLP RR} the two instances of \Cref{alg:rr_sinkhorn} precedently described. 30\% of the values are missing MCAR. All methods are evaluated on 30 random missing values draws. Error bars correspond to $\pm$ 1 std. For readability we display scaled mean $W_2^2$, i.e. for each dataset we renormalize the results by the maximum $W_2^2$. For some datasets $W_2$ results are not displayed due to their large size, which makes evaluating the unregularized $W_2$ distance costly. }
    \label{fig:MCAR_p_3_baselines}
    \vskip-.3cm
\end{figure*}
\textbf{Baselines.}
We compare our methods to three baselines:
\begin{enumerate}[(i)]
    \item \textbf{mean} is the coordinate-wise mean imputation;
    \item \textbf{ice} (imputation by chained equations) consists in (iterative) imputation using conditional expectation. Here, we use \texttt{scikit-learn}'s~\citep{scikit-learn} \texttt{iterativeImputer} method, which is based on \texttt{mice} \citep{buuren2011}. 
    This is one of the most popular methods of imputation as it provides empirically good imputations in many scenario and requires little tuning;
    \item \textbf{softimpute}~\citep{hastie2015softimpute} performs missing values imputation using iterative soft-thresholded SVD's. This method is based on a low-rank assumption for the data and is justified by the fact that many large matrices are well approximated by a low-rank structure \citep{udell2019big}.
\end{enumerate}
\textbf{Deep learning methods.} 
Additionally, we compare our methods to three DL-based methods:
\begin{enumerate}[(i)]
    \setcounter{enumi}{3}
    \item \textbf{MIWAE}~\citep{mattei19a} fits a deep latent variable model (DLVM)~\citep{kingma2014vae}, by optimizing a version of the \emph{importance weighted autoencoder} (IWAE) bound~\citep{burda2016importance} adapted to missing data;
    \item \textbf{GAIN}~\citep{yoon18a} is an adaptation of \emph{generative adversarial networks} (GAN)~\citep{goodfellow2014generative} to missing data imputation;
    \item \textbf{VAEAC}~\citep{ivanov2018variational} are VAEs with easily approximable conditionals that allow to handle missing data.
\end{enumerate}

\textbf{Transport methods.}
Three variants of the proposed methods are evaluated:
\begin{enumerate}[(i)]
    \setcounter{enumi}{6}
    \item \textbf{Sinkhorn} designates the direct non-parametric imputation method detailed in \Cref{alg:sinkhorn_imp}.
\end{enumerate}
For \Cref{alg:rr_sinkhorn}, two classes of imputers are considered: 
\begin{enumerate}[(i)]
    \setcounter{enumi}{7}
    \item \textbf{Linear RR} corresponds to \Cref{alg:rr_sinkhorn} where  for $1\leq j\leq \dim$, $\text{Imputer}(\cdot, \theta_j)$ is a linear model w.r.t. the $\dim-1$ other variables with weights and biases given by $\theta_j$. This is similar to \texttt{mice} or \texttt{IterativeImputer}, but fitted with the OT loss \cref{eq:loss_batches};
\end{enumerate}
\begin{enumerate}[(i)]
    \setcounter{enumi}{8}
    \item \textbf{MLP RR} denotes \Cref{alg:rr_sinkhorn} with shallow Multi-Layer Perceptrons (MLP) as imputers. These MLP's have the following architecture: (i) a first $(\dim-1)\times 2 (\dim-1)$ layer followed by a ReLU layer then (ii) a  $2 (\dim-1)\times (\dim-1)$ layer followed by a ReLU layer and finally (iii) a $(\dim - 1) \times 1$ linear layer. All linear layers have bias terms. Each $\text{Imputer}(\cdot, \theta_j), 1\leq j \leq \dim$ is one such MLP with a different set of weights $\theta_j$.
\end{enumerate}

\textbf{Toy experiments.}
In \Cref{fig:toy_example}, we generate two-dimensional datasets with strong structures, such as an S-shape, half-moon(s), or concentric circles. A 20\% missing rate is introduced (void rows are discarded), and imputations performed using \Cref{alg:sinkhorn_imp} or the \textbf{ice} method are compared to the ground truth dataset.
While the \textbf{ice} method is not able to catch the non-linear structure of the distributions at all, \textbf{Sinkhorn} performs efficiently by imputing faithfully to the underlying complex data structure (despite the two half-moons and the S-shape being quite challenging). This is remarkable, since \Cref{alg:sinkhorn_imp} does not rely on any parametric assumption for the data. This underlines in a low-dimensional setting the flexibility of the proposed method. Finally, note that the trailing points which can be observed for the S shape or the two moons shape come from the fact that \Cref{alg:sinkhorn_imp} was used as it is, i.e. with pairs of batches \emph{both} containing missing values, even though these toy examples would have allowed to use batches without missing values. In that case, we obtain imputations that are visually indistinguishable from the ground truth.
\begin{figure*}[ht]
    \centering
    \includegraphics[width = 1.\textwidth]{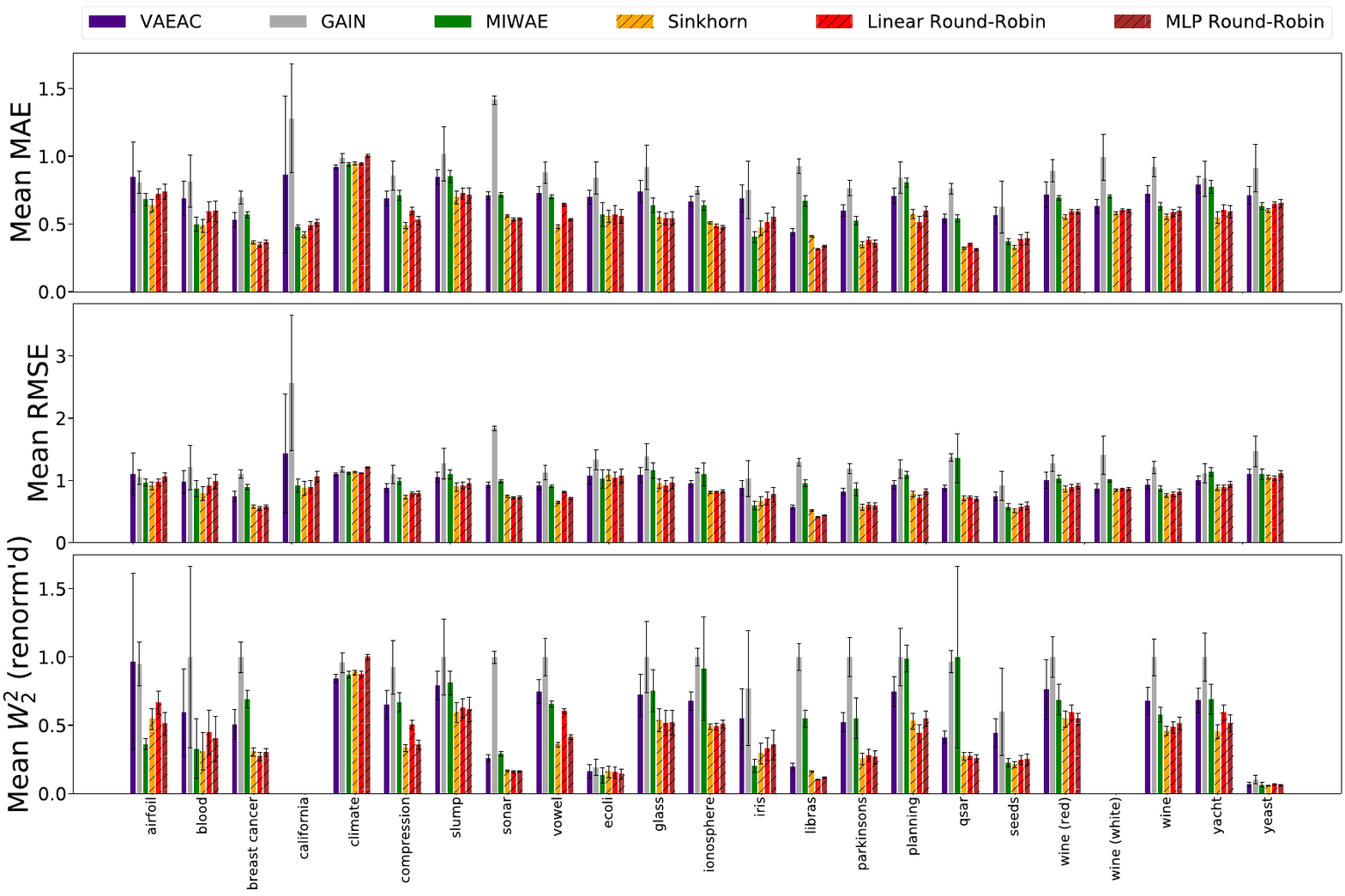}
    \vskip-.3cm
    \caption{(\textbf{30\% MNAR}) Imputation methods on 23 datasets from the UCI repository (\Cref{tab:datasets}). Values are missing MNAR according to the logistic mechanism described in \Cref{sec:exps}, with 30\% variables used as inputs of a logistic masking model for the 70\% remaining variables. 30\% of those input variables are then masked at random. Hence, all variables have 30\% missing values. All methods are evaluated on the same 30 random missing values draws. Error bars correspond to $\pm$ 1 std. For readability we display scaled mean $W_2^2$, i.e. for each dataset we renormalize the results by the maximum $W_2^2$. For some datasets $W_2$ results are not displayed due to their large size, which makes evaluating the unregularized $W_2$ distance costly. }
    \label{fig:MNAR_p_3_DL}
    \vskip-.3cm
\end{figure*}

\textbf{Large-scale experimental setup.}
We evaluate each method on 23 datasets from the UCI machine learning repository\footnote{\url{https://archive.ics.uci.edu/ml/index.php}} (see \Cref{tab:datasets}) with varying proportions of missing data and different missing data mechanisms. These datasets only contain quantitative 
features. Prior to running the experiments, the data is whitened (i.e. centered and scaled to variable-wise unit variance). For each dataset, all methods are evaluated on 30 different draws of missing values masks. For all Sinkhorn-based imputation methods, the regularization parameter $\epsilon$ is set to $5\%$ of the median distance between initialization values with no further dataset-dependent tuning. If the dataset has more than 256 points, the batch size is fixed to 128, otherwise to $2^{\lfloor\frac{n}{2}\rfloor}$ where $n$ is the size of the dataset. The noise parameter $\eta$ in \Cref{alg:sinkhorn_imp} is fixed to $0.1$. For Sinkhorn round-robin models (\textbf{Linear RR} and \textbf{MLP RR}), the maximum number of cycles is 10, 10 pairs of batches are sampled per gradient update, and an $\ell^2$-weight regularization of magnitude $10^{-5}$ is applied during training. For all 3 Sinkhorn-based methods, we use gradient methods with adaptive step sizes as per \cref{alg:sinkhorn_imp,alg:rr_sinkhorn}, with an initial step size fixed to $10^{-2}$. For \textbf{softimpute}, the hyperparameter is selected at each run through cross-validation on a small grid. This CV is performed by sampling additional missing values. For DL-based methods, the implementations provided in open-access by the authors were used\footnote{\url{https://github.com/pamattei/miwae}}\footnote{\url{https://github.com/jsyoon0823/GAIN}}\footnote{\url{https://github.com/tigvarts/vaeac}}, with the hyperparameter settings recommended in the corresponding papers. In particular, for \textbf{GAIN} the $\alpha$ parameter is selected using cross-validation. GPUs are used for Sinkhorn and deep learning methods. The code to reproduce the experiments is available at \url{https://github.com/BorisMuzellec/MissingDataOT}.
\begin{table}[ht]
\centering
\caption{Summary of datasets}\label{tab:datasets}
\vskip-.2cm
\begin{tabular}{lrr}
\toprule
                     dataset &      n &   d \\
\midrule
        airfoil\_self\_noise &   1503 &   5 \\
          blood\_transfusion &    748 &   4 \\
  breast\_cancer\_diagnostic &    569 &  30 \\
                  california &  20640 &   8 \\
     climate\_model\_crashes &    540 &  18 \\
       concrete\_compression &   1030 &   7 \\
             concrete\_slump &    103 &   7 \\
 connectionist\_bench\_sonar &    208 &  60 \\
 connectionist\_bench\_vowel &    990 &  10 \\
                       ecoli &    336 &   7 \\
                       glass &    214 &   9 \\
                  ionosphere &    351 &  34 \\
                        iris &    150 &   4 \\
                      libras &    360 &  90 \\
                  parkinsons &    195 &  23 \\
             planning\_relax &    182 &  12 \\
        qsar\_biodegradation &   1055 &  41 \\
                       seeds &    210 &   7 \\
                        wine &    178 &  13 \\
          wine\_quality\_red &   1599 &  10 \\
        wine\_quality\_white &   4898 &  11 \\
        yacht\_hydrodynamics &    308 &   6 \\
                       yeast &   1484 &   8 \\
\bottomrule
\end{tabular}
\vskip-.4cm
\end{table}

\textbf{Missing value generation mechanisms.}
The implementation of a MCAR mechanism is straightfoward. On the contrary, many different mechanisms can lead to a MAR or MNAR setting. We here describe those used in our experiments. In the \textbf{MCAR} setting, each value is masked according to the realization of a Bernoulli random variable with a fixed parameter. In the \textbf{MAR} setting, for each experiment, a fixed subset of variables that cannot have missing values is sampled. Then, the remaining variables have missing values according to a logistic model with random weights, which takes the non-missing variables as inputs. A bias term is fitted using line search to attain the desired proportion of missing values. Finally, two different mechanisms are implemented in the \textbf{MNAR} setting. The first is identical to the previously described MAR mechanism, but the inputs of the logistic model are then masked by a MCAR mechanism. Hence, the logistic model's outcome now depends on potentially missing values. The second mechanism, 'self masked',  samples a subset of variables whose values in the lower and upper $p$-th percentiles are masked according to a Bernoulli random variable, and the values in-between are left not missing. As detailed in the appendix, MCAR experiments were performed with 10\%, 30\% and 50\% missing rates, while MAR and both MNAR settings (quantile and logistic masking) were evaluated with a 30\% missing rate.

\textbf{Metrics.}
Imputation methods are evaluated according to two ``pointwise'' metrics: mean absolute error (MAE) and root mean square error (RMSE); and one metric on distributions: the squared Wasserstein distance between empirical distributions on points with missing values. 
Let $\bX \in \RR^{n\times \dim}$ be a dataset with missing values. 
When $(i,j)$ spots a missing entry, recall that $\hat{x}_{ij}$ denotes the corresponding imputation, and let us note $x^{\text{true}}_{ij}$ the ground truth. 
Let $m_0 \defeq \#\{(i, j) , \omega_{ij}=0 \}$  and $m_1 \defeq \#\{i : \exists j , \, \,  \omega_{ij}=0 \}\}$ respectively denote the total number of missing values and the number of data points with at least one missing value. Set $M_1 \defeq \{i : \exists j , \omega_{ij}=0 \}$. We define MAE, RMSE and $W_2$ imputation metrics as
\begin{align*}\label{eq:metrics}
    \tfrac{1}{m_0} \sum_{\mathclap{(i, j) : \omega_{ij} = 0}} |x_{i,j}^{\text{true}} - \hat{x}_{ij}|,\tag{MAE}\\
    \sqrt{\tfrac{1}{m_0}\sum_{\mathclap{(i, j) : \omega_{ij} = 0}}  (x_{i,j}^{\text{true}} - \hat{x}_{ij})^2},\tag{RMSE}\\
     W_2^2\left( \mu_{m_1}(\hat{\data}_{M_1}),\mu_{m_1}(\data^{(true)}_{M_1})\right).\tag{$W_2$}
\end{align*}

\textbf{Results.}
The complete results of the experiments are reported in the Appendix. In \Cref{fig:MCAR_p_3_baselines} and \Cref{fig:MNAR_p_3_DL}, the proposed methods are respectively compared to baselines and Deep Learning (DL) methods in a MCAR and a logistic masking MNAR setting with 30\% missing data. As can be seen from \Cref{fig:MCAR_p_3_baselines}, the linear round-robin model matches or out-performs \texttt{scikit}'s iterative imputer (\textbf{ice}) on MAE and RMSE scores for most datasets. Since both methods are based on the same cyclical linear imputation model but with different loss functions, this shows that the batched Sinkhorn loss in \Cref{eq:loss_batches} is well-adapted to imputation with parametric models. Comparison with DL methods (\Cref{fig:MNAR_p_3_DL}) shows that the proposed OT-based methods consistently outperform DL-based methods, and have the additional benefit of having a lower variance in their results overall. Interestingly, while the MAE and RMSE scores of the round-robin MLP model are comparable to that of the linear RR, its $W_2$ scores are generally better. This suggests that more powerful base imputer models lead to better $W_2$ scores, from which one can conclude that \Cref{eq:loss_batches} is a good proxy for optimizing the unavailable \Cref{eq:W2} score, and that Algorithm \ref{alg:rr_sinkhorn} is efficient at doing so. Furthermore, one can observe that the direct imputation method is very competitive over all data and metrics and is in general the best performing OT-based method, as could be expected from the fact that its imputation model is not restricted by a parametric assumption. This favorable behaviour tends to be exacerbated with a growing proportion of missing data, see \Cref{fig:MCAR_p_5_summary} in the appendix.

\textbf{MAR and MNAR.}
\Cref{fig:MNAR_p_3_DL} above and \Cref{fig:MAR_p_3_summary,fig:MNAR_log_p_3_summary_2,fig:MNAR_quant_p_3_summary} in the appendix display the results of our experiments in MAR and MNAR settings, and show that the proposed methods perform well and are robust to difficult missingness mechanisms. This is remarkable, as the proposed methods do not attempt to model those mechanisms. Finally, note that the few datasets on which the proposed methods do not perform as well as baselines -- namely \texttt{libras} and to a smaller extent \texttt{planning\_relax} -- remain consistently the same across all missingness mechanisms and missing rates. This suggests that this behavior is due to the particular structure of those datasets, rather than to the missingness mechanisms themselves.

\textbf{Out-of-sample imputation.}
\begin{figure}
    \centering
    \includegraphics[width = 1.\linewidth]{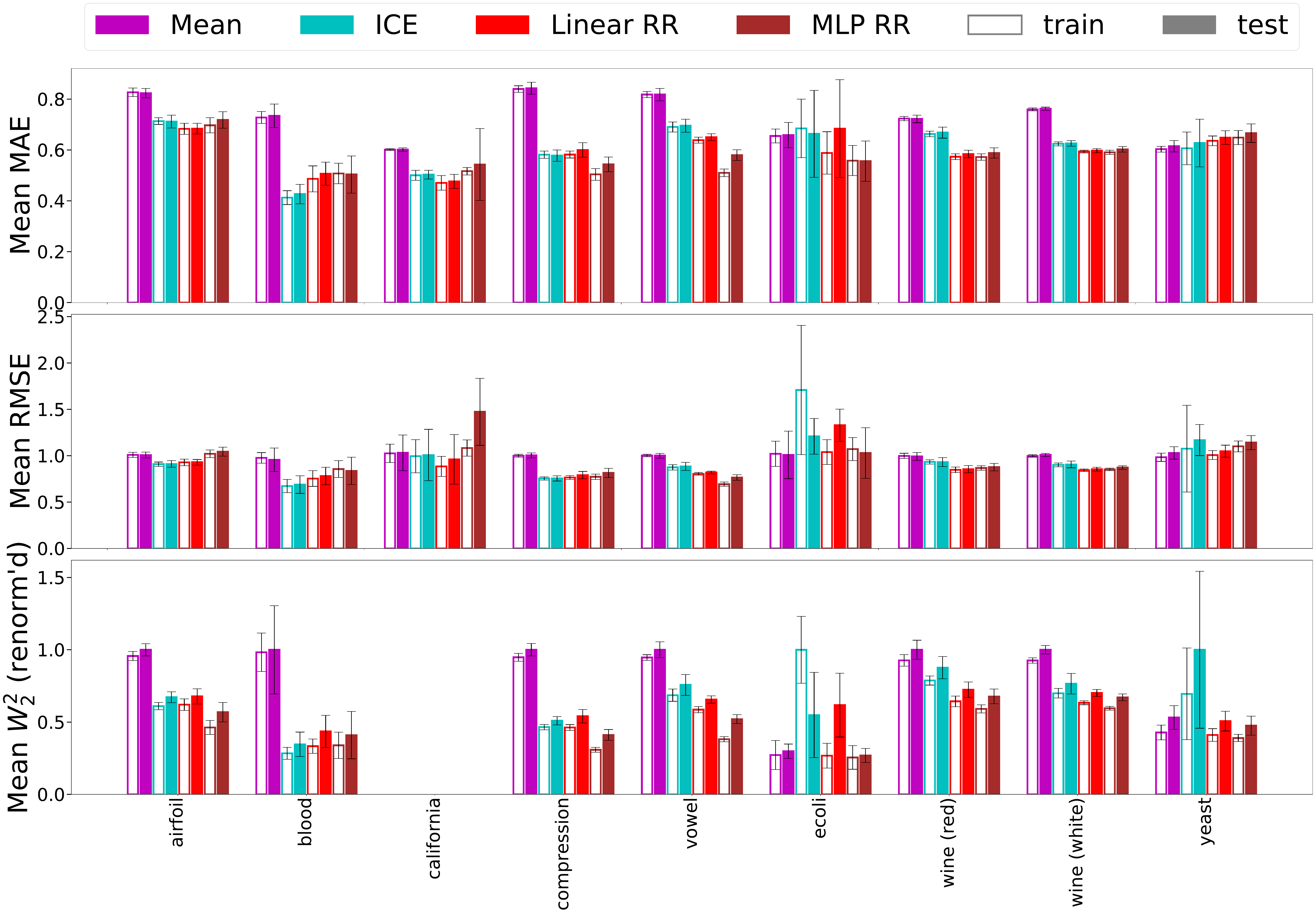}
    \vskip-.2cm
    \caption{(\textbf{OOS}) Out of sample imputation: 70\% of the data is used for training (filled bars) and 30 \% for testing with fixed parameters (dotted bars). 30\% of the values are missing MCAR accross both training and testing sets.}
    \label{fig:OOF_p_3_summary}
    \vskip-.5cm
\end{figure}
As mentioned in \Cref{sec:miss_data_ot}, a key benefit of fitting a parametric imputing model with \cref{alg:meta_sinkhorn,alg:rr_sinkhorn} is that the resulting model can then be used to impute missing values in out-of-sample (OOS) data. In \Cref{fig:OOF_p_3_summary}, we evaluate the Linear RR and MLP RR models in an OOS imputation experiment. We compare the training and OOS MAE, RMSE and OT scores on a collection of datasets selected to have a sufficient number of points. At each run, we randomly sample 70\% of the data to be used for training, and the remaining 30\% to evaluate OOS imputation. 30\% of the values are missing MCAR, uniformly over training and testing sets. Out of the methods presented earlier on, we keep those that allow OOS: for the \textbf{ice}, \textbf{Linear RR} and \textbf{MLP RR} methods, OOS imputation is simply performed using the round-robin scheme without further fitting of the parameters on the new data. For the \textbf{mean} baseline, missing values in the testing data are imputed using mean observed values from the training data. \Cref{fig:OOF_p_3_summary} confirms the stability at testing time of the good performance of \textbf{Linear RR} and \textbf{MLP RR}.

\section*{Conclusion}
We have shown in this paper how OT metrics could be used to define a relevant loss for missing data imputation. This loss corresponds to the expectation of Sinkhorn divergences between randomly sampled batches. To minimize it, two classes of algorithms were proposed: one that freely estimates one parameter per imputed value, and one that fits a parametric model. The former class does not rely on making parametric assumptions on the underlying data distribution, and can be used in a very wide range of settings. On the other hand, after training, the latter class allows out-of-sample imputation. To make parametric models trainable, the classical round-robin mechanism was used. Experiments on a variety of datasets, and for numerous missing value settings (MCAR, MAR and MNAR with varying missing values proportions) showed that the proposed models are very competitive, even compared to recent methods based on deep learning. These results confirmed that our loss is a good optimizable proxy for imputation metrics. 
Future work includes further theoretical study of our loss function \Cref{eq:loss_batches} within the OT framework.

\clearpage

\bibliography{references}
\bibliographystyle{icml2020}


\clearpage
\appendix
\section{Appendix}

\begin{figure*}[ht]
  \begin{subfigure}{.95\textwidth}
      \centering
      \includegraphics[width = 1.\textwidth]{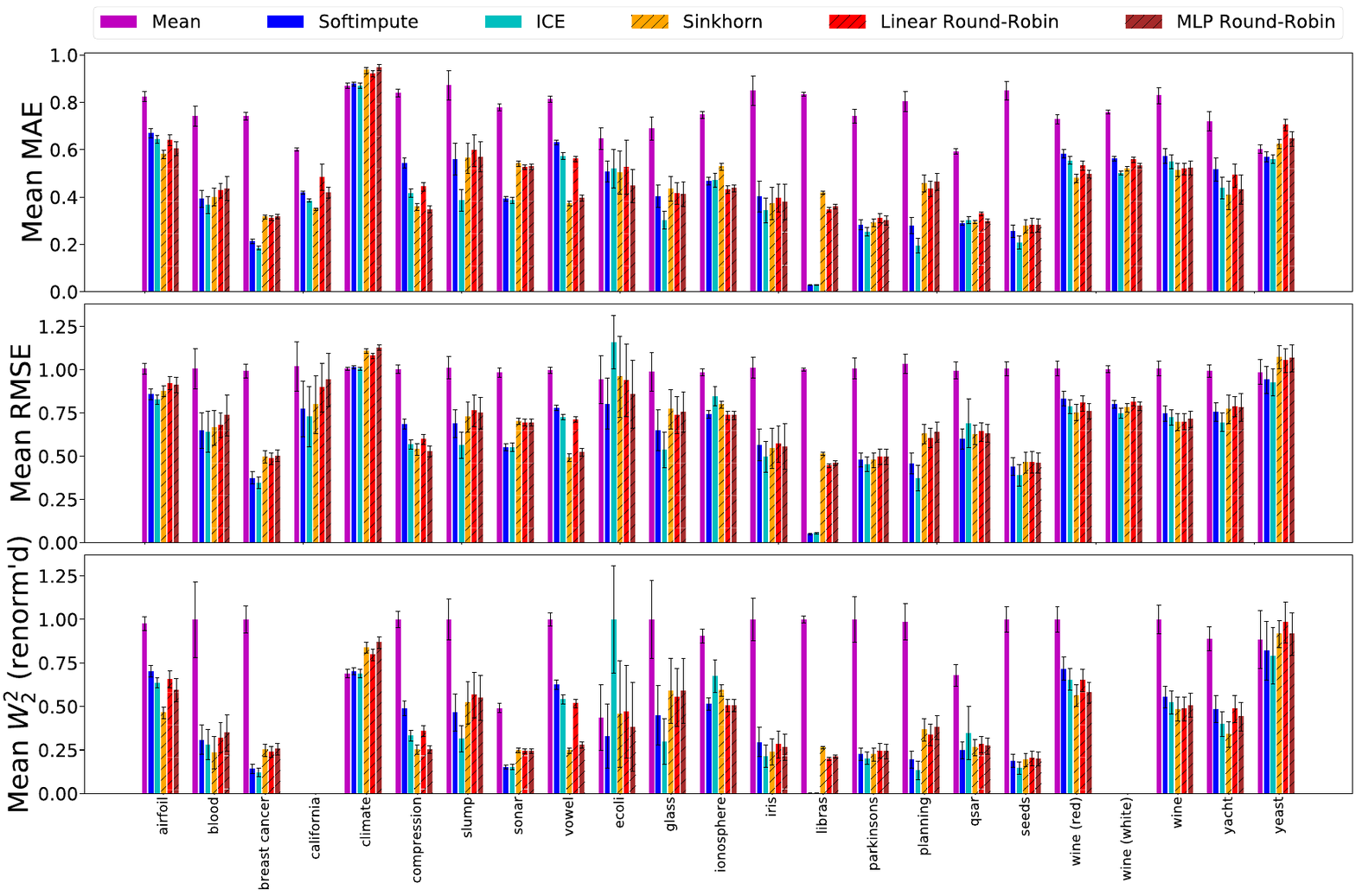}
  \end{subfigure}
  \vskip-.8cm
    \begin{subfigure}{.95\textwidth}
      \centering
      \includegraphics[width = 1.\textwidth]{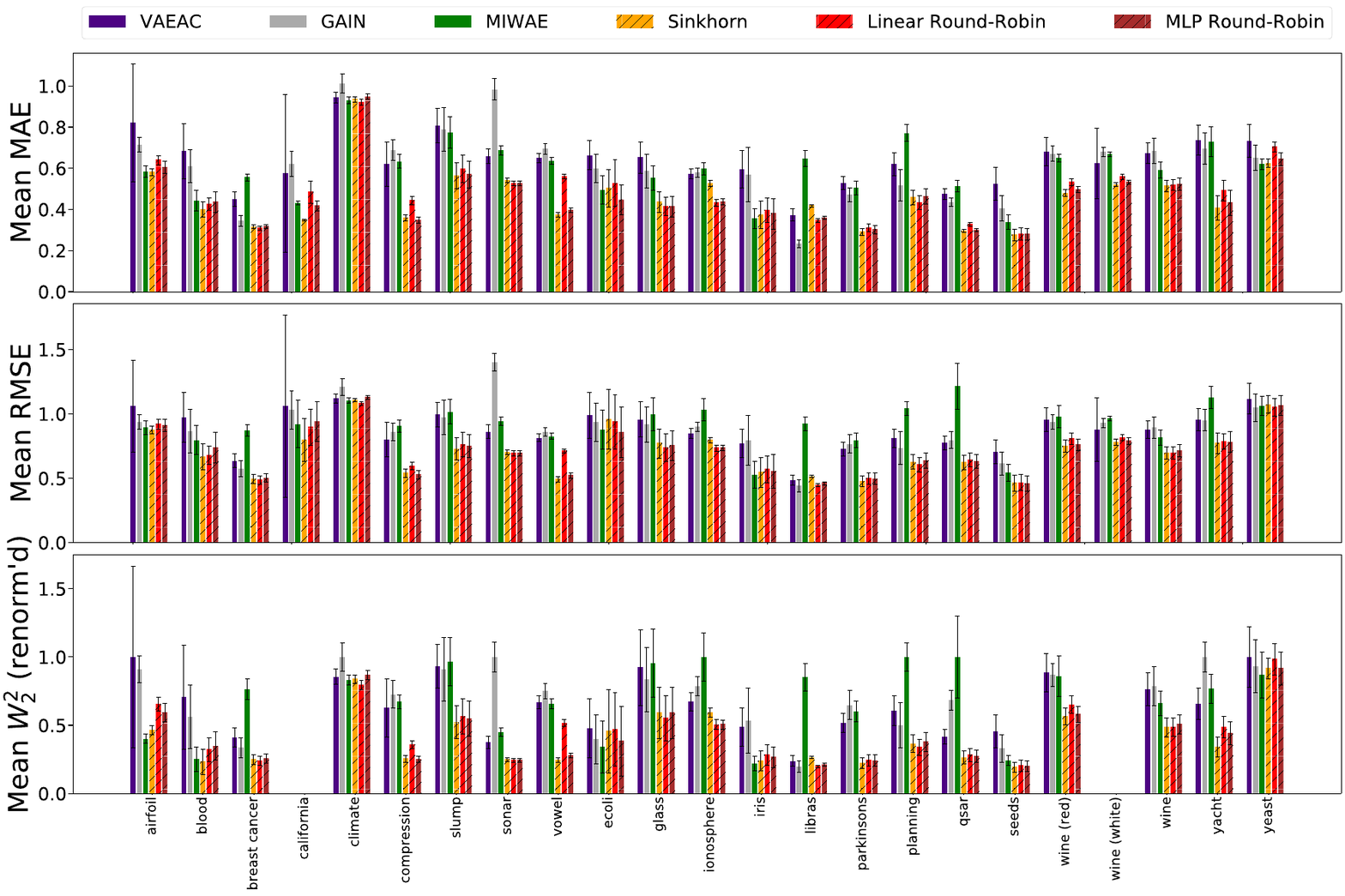}
  \end{subfigure}
    \vskip-.5cm
    \caption{(\textbf{10 \% MCAR})}
    \label{fig:MCAR_p_1_summary}
\end{figure*}

\begin{figure*}[ht]
  \begin{subfigure}{.95\textwidth}
      \centering
      \includegraphics[width = 1.\textwidth]{figs/baselines_summary_mean_error_dataset_GAN_p_3.pdf}
  \end{subfigure}
  \vskip-.8cm
    \begin{subfigure}{.95\textwidth}
      \centering
      \includegraphics[width = 1.\textwidth]{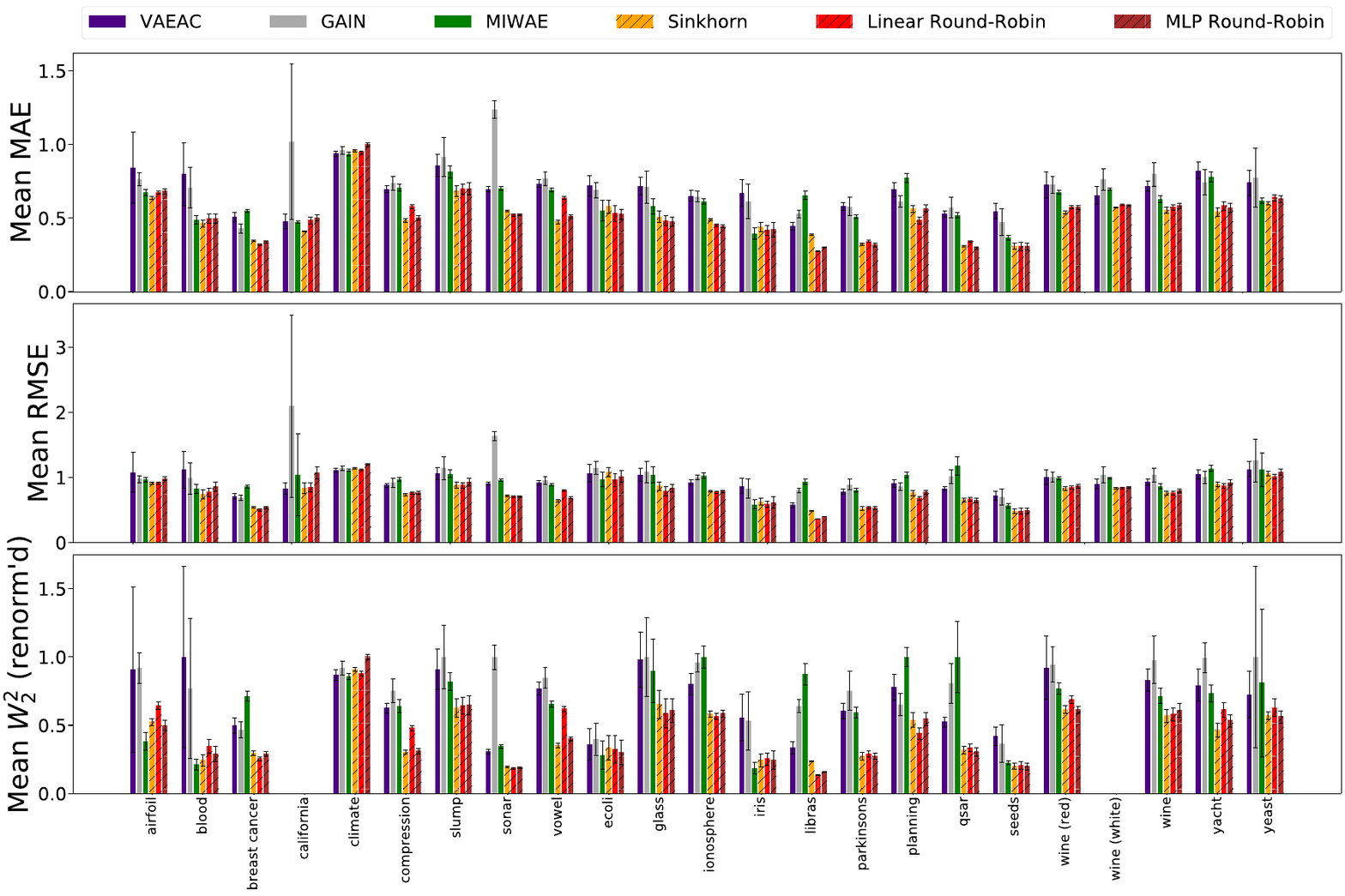}
  \end{subfigure}
    \vskip-.5cm
    \caption{(\textbf{30 \% MCAR})}
    \label{fig:MCAR_p_3_summary}
\end{figure*}

\begin{figure*}[ht]
  \begin{subfigure}{.95\textwidth}
      \centering
      \includegraphics[width = 1.\textwidth]{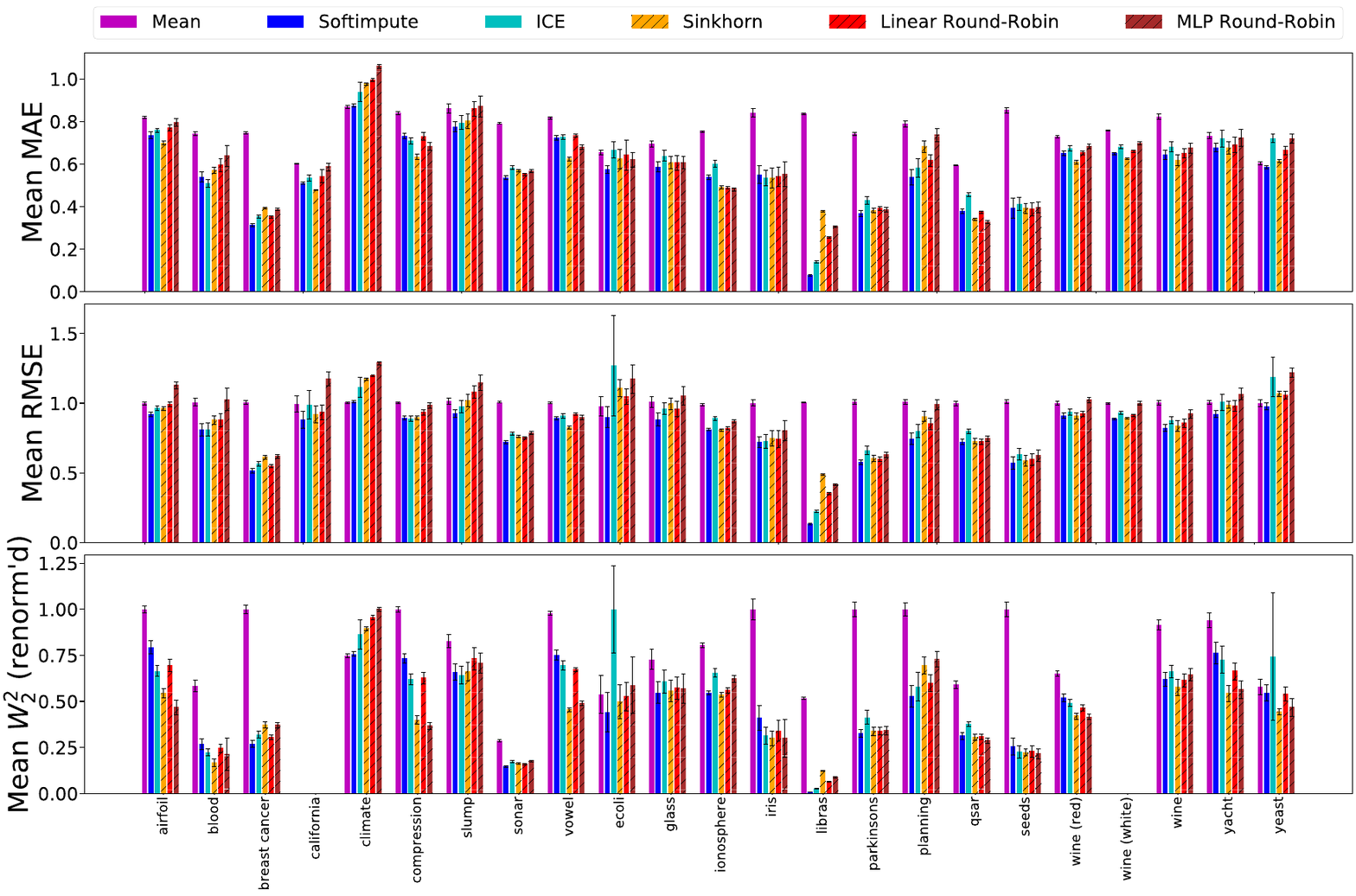}
  \end{subfigure}
  \vskip-.8cm
    \begin{subfigure}{.95\textwidth}
      \centering
      \includegraphics[width = 1.\textwidth]{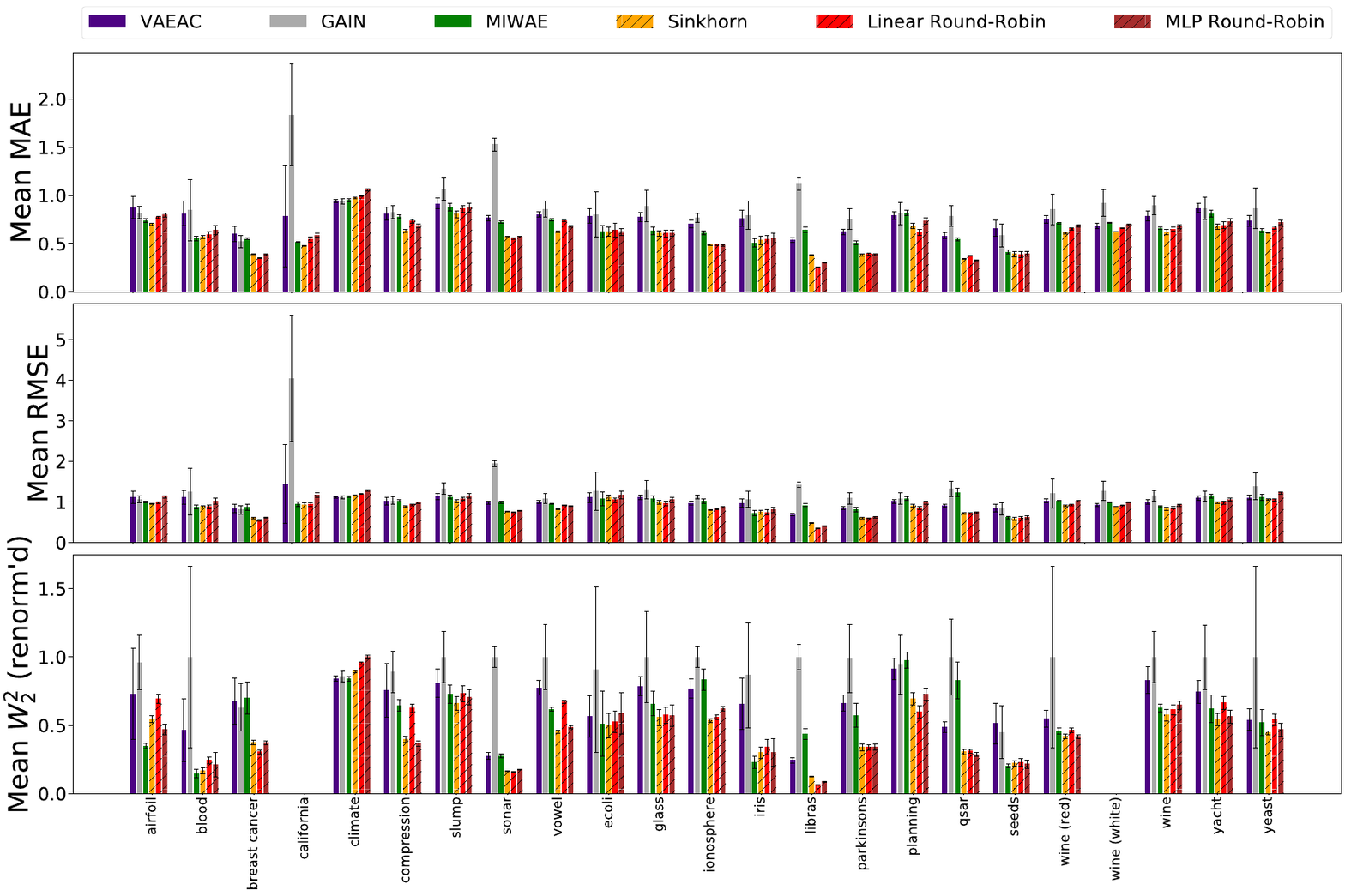}
  \end{subfigure}
    \vskip-.5cm
    \caption{(\textbf{50 \% MCAR})}
    \label{fig:MCAR_p_5_summary}
\end{figure*}

\begin{figure*}[ht]
  \begin{subfigure}{.95\textwidth}
      \centering
      \includegraphics[width = 1.\textwidth]{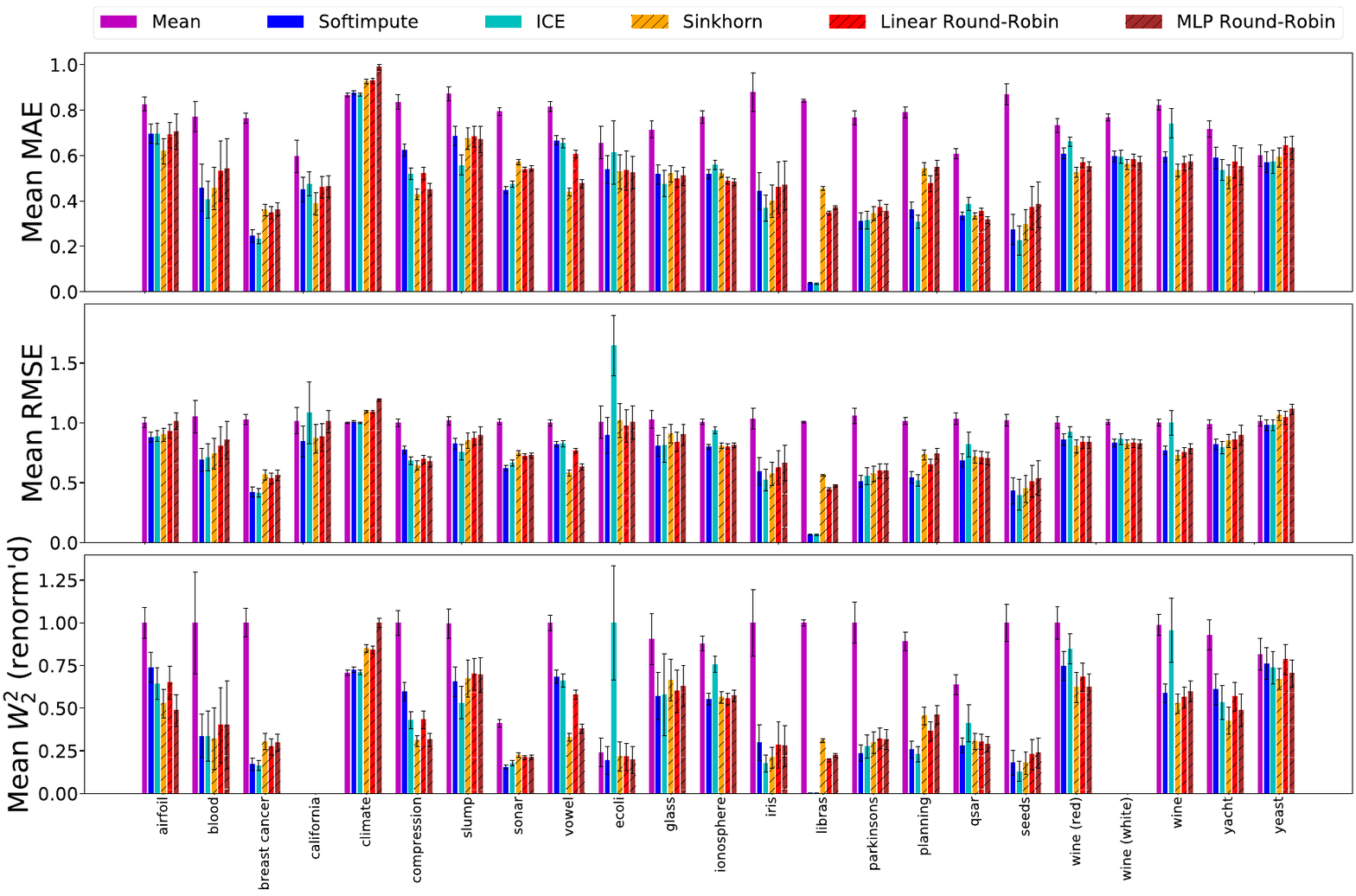}
  \end{subfigure}
  \vskip-.8cm
    \begin{subfigure}{.95\textwidth}
      \centering
      \includegraphics[width = 1.\textwidth]{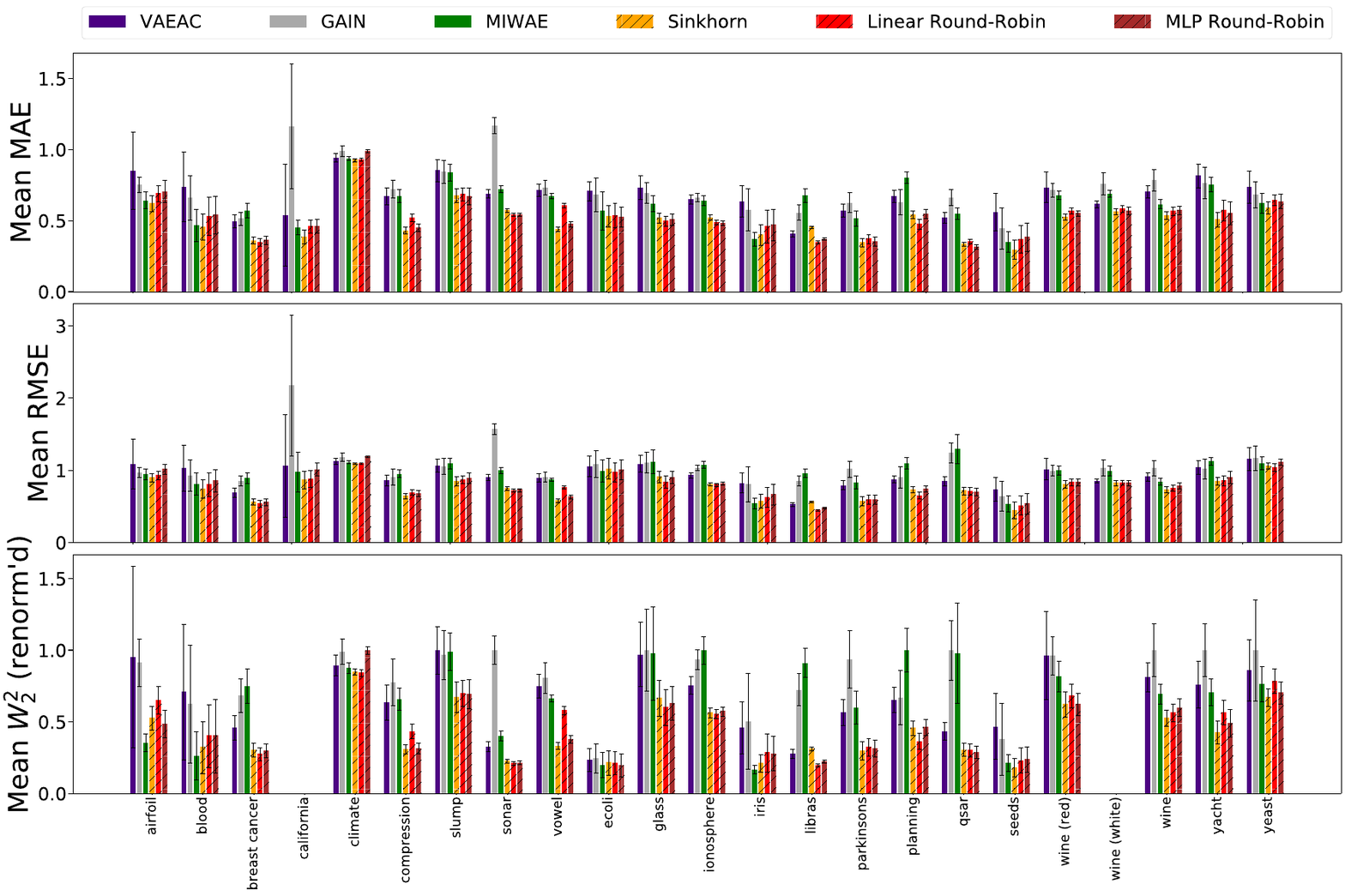}
  \end{subfigure}
    \vskip-.5cm
    \caption{(\textbf{30 \% MAR})}
    \label{fig:MAR_p_3_summary}
\end{figure*}

\begin{figure*}[ht]
  \begin{subfigure}{.95\textwidth}
      \centering
      \includegraphics[width = 1.\textwidth]{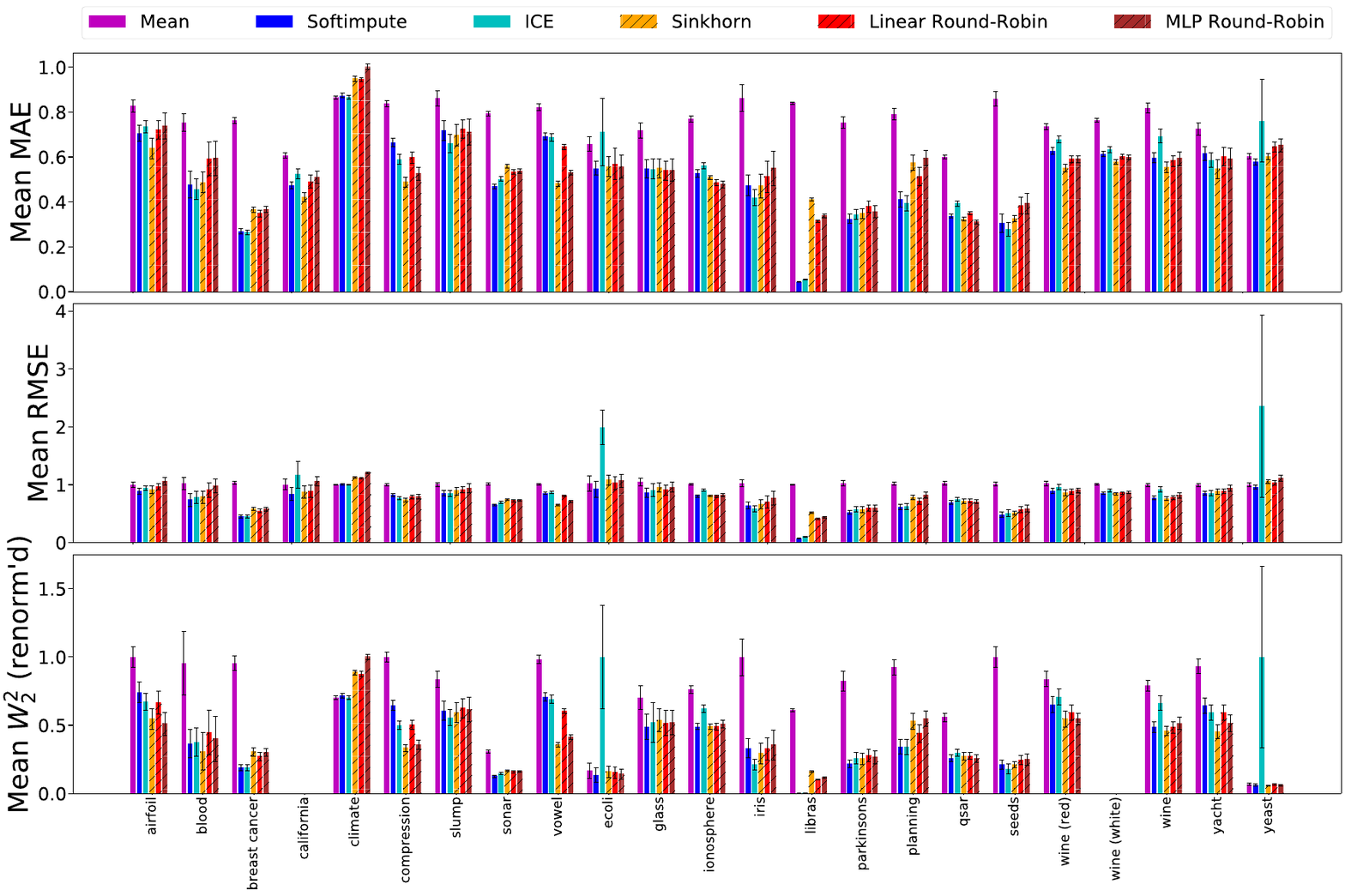}
  \end{subfigure}
  \vskip-.8cm
    \begin{subfigure}{.95\textwidth}
      \centering
      \includegraphics[width = 1.\textwidth]{figs/DL_summary_mean_error_dataset_MNAR_log_fix_p_3.pdf}
  \end{subfigure}
    \vskip-.5cm
    \caption{(\textbf{30 \% MNAR, logistic masking})}
    \label{fig:MNAR_log_p_3_summary_2}
\end{figure*}

\begin{figure*}[ht]
  \begin{subfigure}{.95\textwidth}
      \centering
      \includegraphics[width = 1.\textwidth]{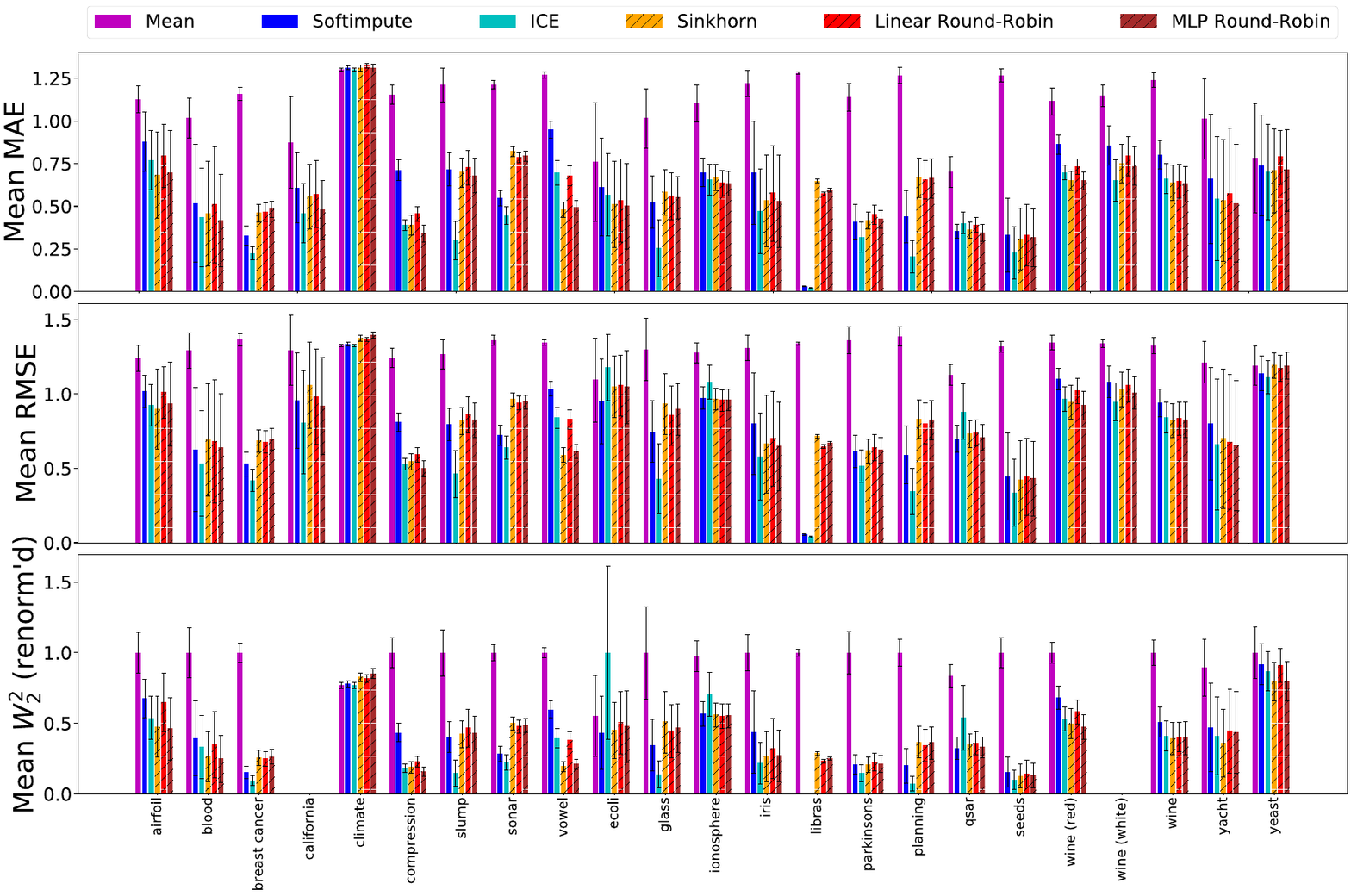}
  \end{subfigure}
  \vskip-.8cm
    \begin{subfigure}{.95\textwidth}
      \centering
      \includegraphics[width = 1.\textwidth]{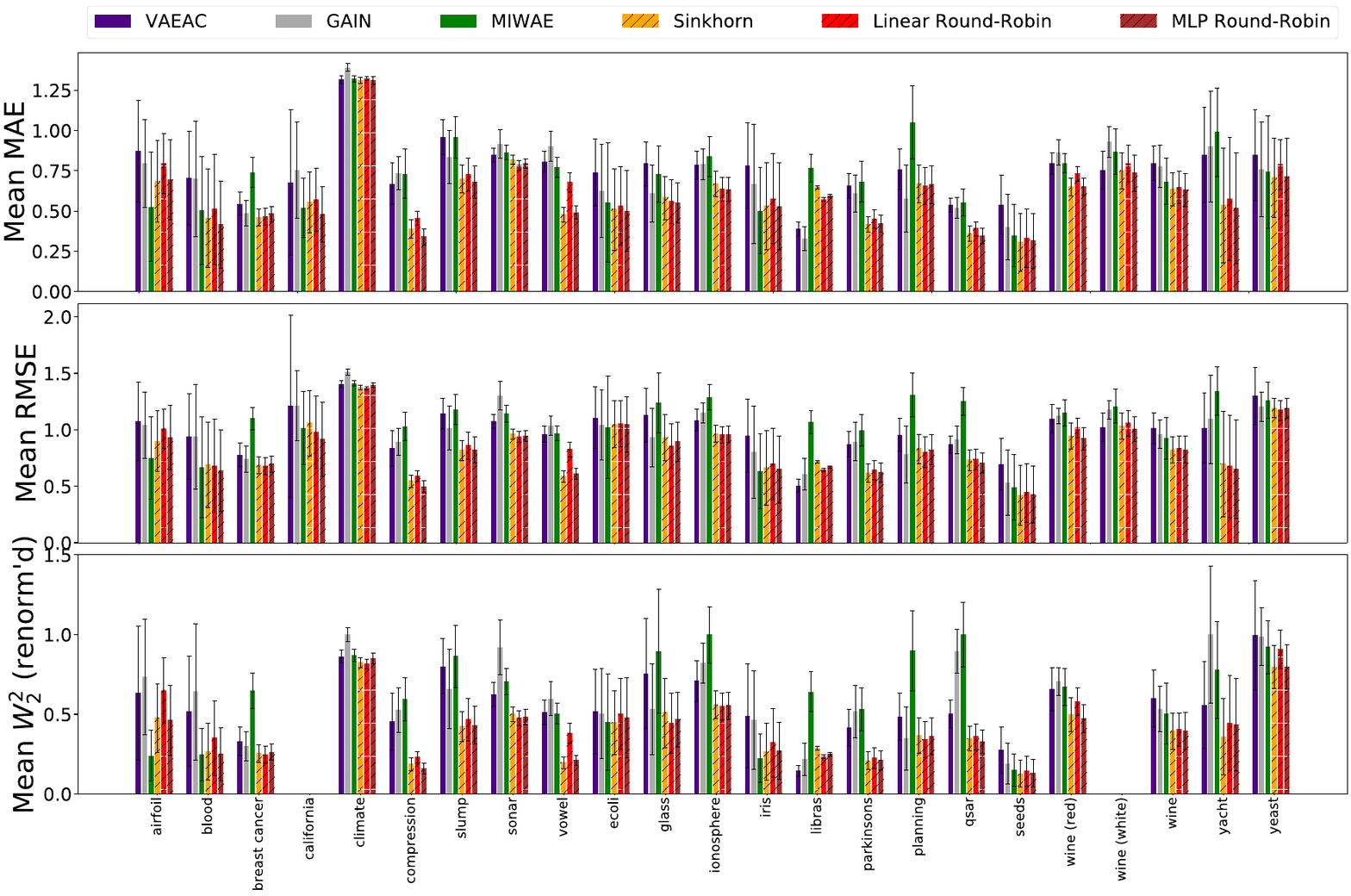}
  \end{subfigure}
    \vskip-.5cm
    \caption{(\textbf{30 \% MNAR, quantile masking})}
    \label{fig:MNAR_quant_p_3_summary}
\end{figure*}

This appendix contains a full account of our experimental results. These results correspond to the missing value mechanisms described in \Cref{sec:exps}:

\begin{enumerate}
    \item 10\% MCAR (\Cref{fig:MCAR_p_1_summary}), 30\% MCAR (\Cref{fig:MCAR_p_3_summary}) and 50\% MCAR  (\Cref{fig:MCAR_p_5_summary});
    \item 30\% MAR on 70\% of the variables with a logistic masking model (\Cref{fig:MAR_p_3_summary});
    \item 30\% MNAR generated with a logistic masking model, whose inputs are then themselves masked (\Cref{fig:MNAR_log_p_3_summary_2});
    \item 30\% MNAR on 30\% of the variables, generated by censoring upper and lower quartiles (\Cref{fig:MNAR_quant_p_3_summary}).
\end{enumerate}

These experiments follow the setup described in \Cref{sec:exps}. In all the following figures, error bars correspond to $\pm 1$ standard deviation across the $30$ runs performed on each dataset. For some datasets, the $W_2$ score is not represented: this is due to their large size, which makes computing unregularized OT computationally intensive. 

The results show that the proposed methods, \Cref{alg:sinkhorn_imp} and \Cref{alg:rr_sinkhorn} with linear and shallow MLP imputers, are very competitive compared to state-of-the-art methods, including those based on deep learning~\citep{mattei19a,yoon18a,ivanov2018variational}, in a wide range of missing data regimes.

\paragraph{Runtimes.}

\begin{figure}[ht]
    \centering
    \includegraphics[width = 1.\columnwidth]{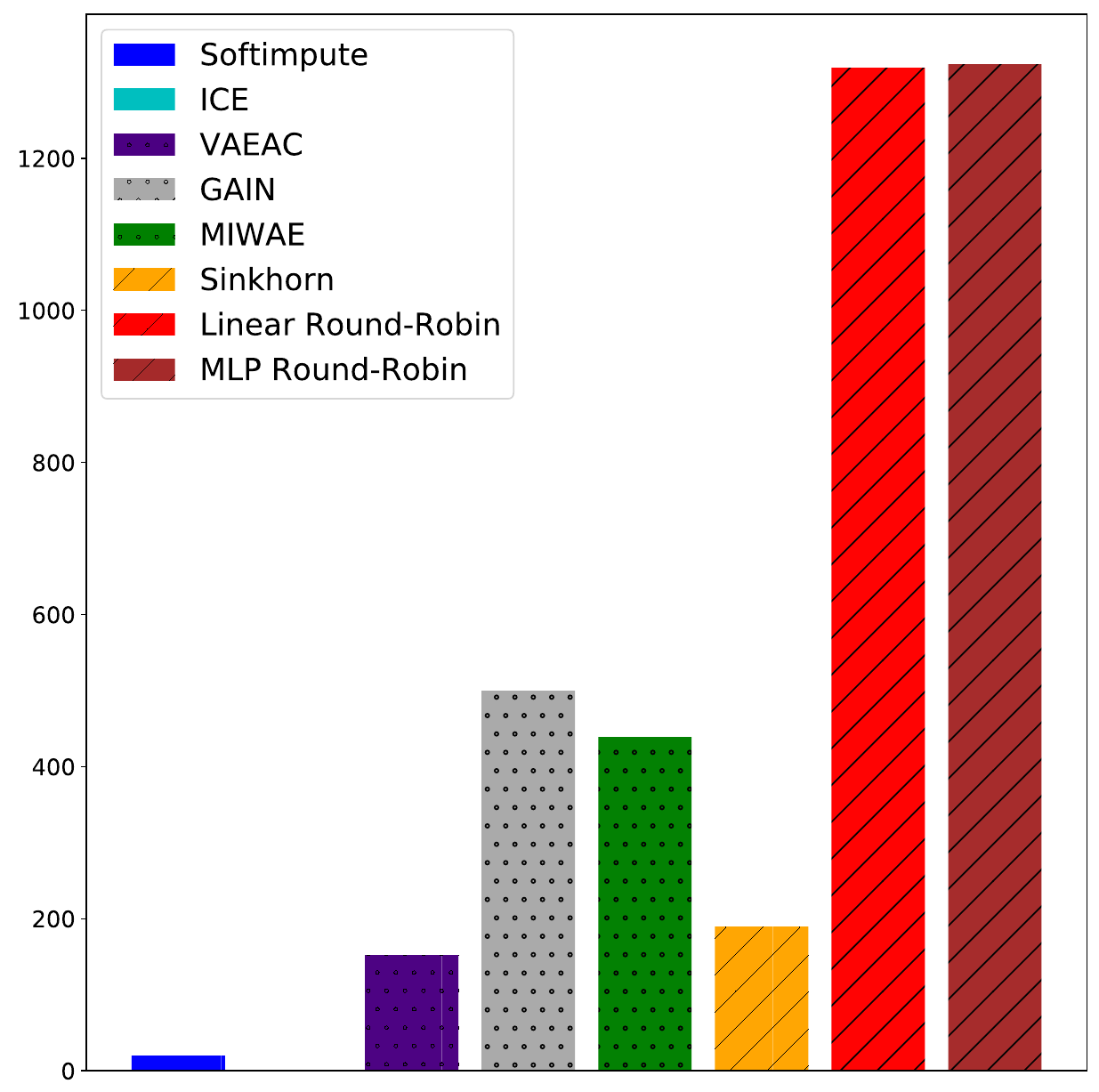}
    \caption{Average runtimes (in seconds, over 30 runs and 23 datasets) for the experiment described in \cref{fig:MNAR_log_p_3_summary_2}. Note that these times are indicative, as runs where randomly assigned to different GPU models, which may have an impact on runtimes.}
    \label{fig:runtimes}
\end{figure}

\Cref{fig:runtimes} represents the average runtimes of the methods evaluated in \Cref{fig:MNAR_log_p_3_summary_2}. These runtimes show that \Cref{alg:sinkhorn_imp} has computational running times on par with VAEAC, and faster than the two remaining DL-based methods (GAIN and MIWAE). Round-robin methods are the slowest overall, but the base imputer model being used seems to have nearly no impact on runtimes. This is due to the fact that the computational bottleneck of the proposed methods is the number of Sinkhorn batch divergences that are computed. This number can be made lower by e.g. reducing the number of gradient steps performed for each variable (parameter $K$
in \cref{alg:rr_sinkhorn}), or the number of cycles $t_{max}$. This fact suggests that more complex models could be used in round-robin imputation without much additional computational cost. 

\end{document}